# Person Re-Identification using Deep Learning Networks: A Systematic Review

Ankit Yadav[1], Dinesh Kumar Vishwakarma[2]

Biometric Research Laboratory, Department of Information Technology, Delhi Technological University, Bawana Road, Delhi-110042, India

ankit4607@gmail.com[1], dinesh@dtu.ac.in[2]

**Abstract:** Person re-identification has received a lot of attention from the research community in recent times. Due to its vital role in security based applications, person re-identification lies at the heart of research relevant to tracking robberies, preventing terrorist attacks and other security critical events. While the last decade has seen tremendous growth in re-id approaches, very little review literature exists to comprehend and summarize this progress. This review deals with the latest state-of-the-art deep learning based approaches for person re-identification. While the few existing re-id review works have analysed re-id techniques from a singular aspect, this review evaluates numerous re-id techniques from multiple deep learning aspects such as deep architecture types, common Re-Id challenges (variation in pose, lightning, view, scale, partial or complete occlusion, background clutter), multi-modal Re-Id, cross-domain Re-Id challenges, metric learning approaches and video Re-Id contributions. This review also includes several re-id benchmarks collected over the years, describing their characteristics, specifications and top re-id results obtained on them. The inclusion of the latest deep re-id works makes this a significant contribution to the re-id literature. Lastly, the conclusion and future directions are included.

## 1. Introduction

Security and surveillance based applications in computer vision have gained tremendous popularity in recent times. Currently, security surveillances record videos and images whose analysis requires manual human interaction. Looking into yesterday's robbery can be challenging as it involves a manual search through twenty hours of surveillance videos by humans prone to fatigue and making errors. The problem quickly becomes infeasible as the time span of recorded media under analysis is increased. The development of machine learning and later, deep learning approaches have opened a wide range of advanced possibilities that could lead to safer homes, offices, neighbourhood, bus stops, airports etc. The idea that machines can be taught to identify individuals of interest is a promising step towards a more secure environment.

Person re-identification means to find a person of interest in a query image/video from a large collection of recorded images and/or videos. While machine learning algorithms played crucial part in the early days, re-id approaches have made significant improvements with the rise of deep learning based systems [1]. Several deep learning based approaches proposed in recent years have boosted the matching accuracy results, significantly outperforming the handcrafted feature based machine learning algorithms [2], [3]. Since deep learning models require a large number of training samples, recent years have also witnessed the collection of several medium to large-scale re-id datasets for training and testing different deep based re-id approaches.

## 2. Research Methodology

This section details the major contributions of this review, techniques followed in preparing this review and a comparison with some of the existing deep Re-Id reviews/surveys.

### 2.1 Contributions of this Review

This review studies deep learning based person re-identification. While person re-identification is not a new topic in the research community, deep learning based methods have become increasingly popular due their tremendous success in various computer vision domains. Hence, it is natural that deep learning based methods take lead in the Re-Id problem due to their superior feature learning capability when compared to hand-crafted feature based methods.

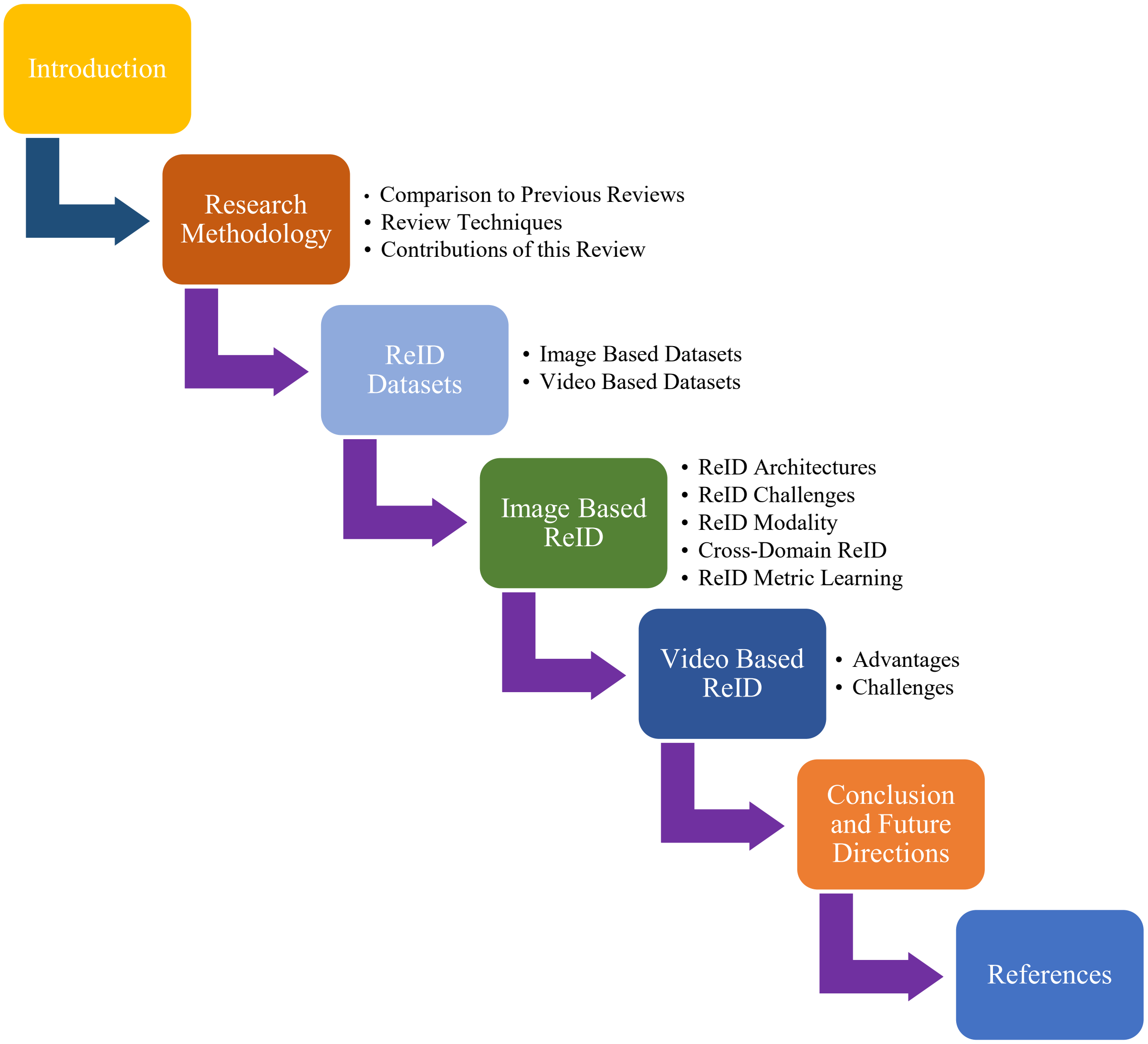

Fig. 1 Organization of this Review

The focus of this review is to conduct an exhaustive study of the recent deep learning based Re-Id methods, specifying the various benchmark datasets and the classification of various image and video based deep methods as shown in Fig. 1, describing the organization of this review. Recent years have witnessed a sharp increase in the number of deep learning based Re-Id approaches and this review cites the latest developments in deep learning based Re-Id research. Fig. 2 describes the taxonomy of deep learning based Re-Id methods included in this review. Re-Id methods have been categorized into image-based and video-based methods. Image-based contributions have been explored from numerous aspects such as architectures involved, common visual challenges, modality specific methods, cross-domain approaches and metric learning methods.

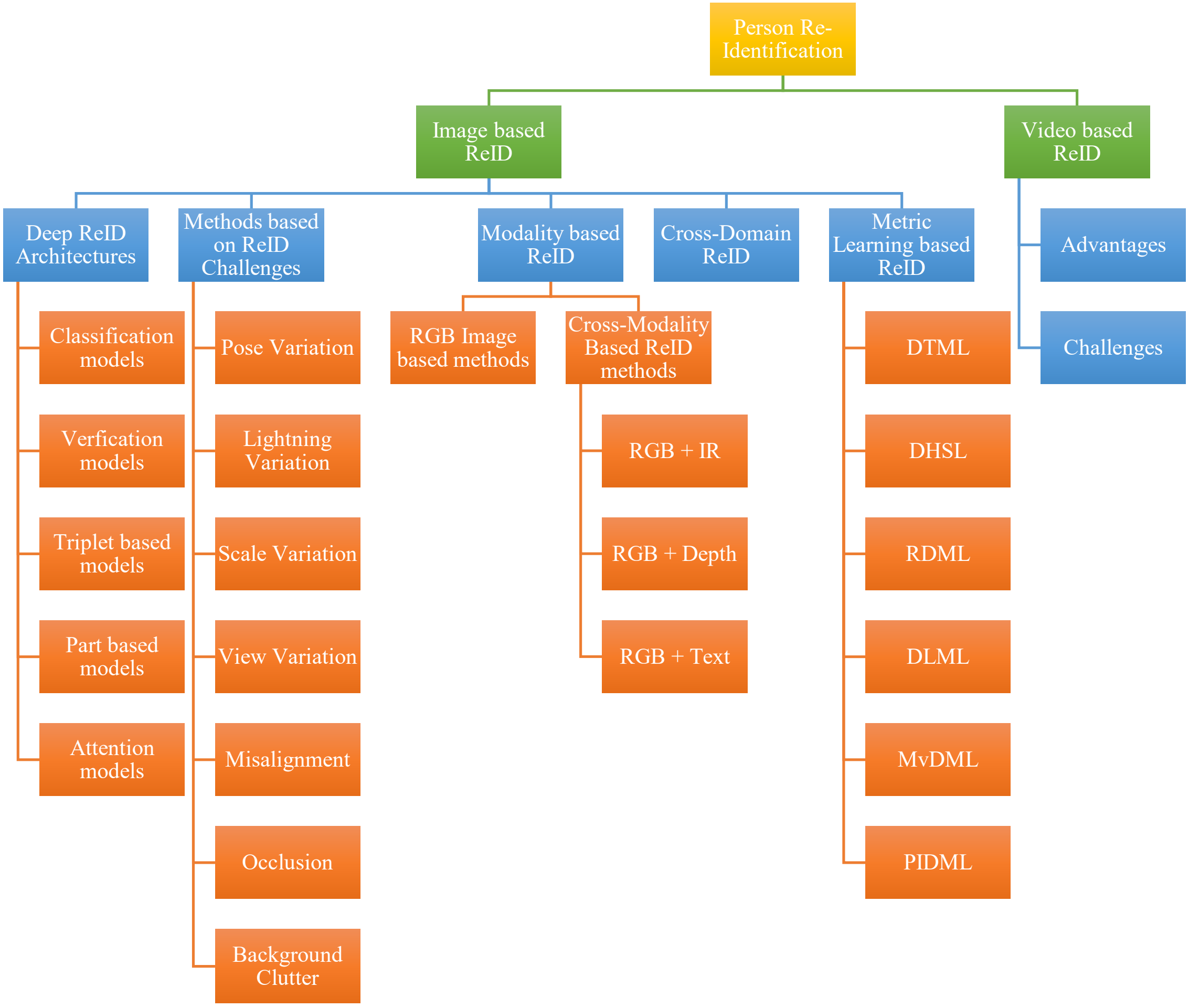

Fig. 2 Taxonomy of Re-Id methods

The major contributions of this review are as follows:

- Provides a comprehensive review of deep learning based person re-identification methods.
- Conducts an exhaustive study of deep Re-Id methods (Table 3-14) by describing the "key ideas" behind numerous approaches mentioned.
- Since deep learning methods have gained popularity in recent years, this exhaustive review automatically incorporates the most recent contributions to deep learning based Re-Id approaches.

- Details several image based and video based benchmark datasets, specifying their technical specifications, challenges posed by samples and top results reported on them.
- Analyses deep learning based methods from several crucial aspects like architecture types, loss functions, Re-Id challenges, data modalities, cross-domain approaches and metric learning methods, helping the readers to understand and appreciate deep Re-Id from multiple perspectives.
- Explores the growing popularity of video based deep Re-Id methods that combine temporal data providing important motion cues with the usual visual characteristics.
- This review helps readers to gain a comprehensive and exhaustive understanding of the recent deep Re-Id contributions by categorically analysing contributions from numerous perspectives like architecture, challenges, modality, cross-domain approaches, metric learning and video based methods.

### 2.2 Review Techniques

This review includes journal papers, conference and workshop papers from several well-known repositories including IEEE Xplore, ScienceDirect, Springer, ACM and Google Scholar. The keywords used to search for relevant contributions include “person re-identification”, “Re-Id”, “deep”, “deep learning”, “review” and “survey”. Due to the growing popularity of deep approaches, this initial search resulted in a comprehensive list of contributions. Higher priority was given to publications from high quality journals such as IEEE Transactions, Pattern Recognition, Neurocomputing etc and top conferences such as CVPR, ECCV and ICCV. A separate search was conducted to include Re-Id dataset contributions. Finally, the included papers were analysed and a taxonomy of deep Re-Id methods was formulated as demonstrated in Fig. 2. Fig. 3 shows the graph of year-wise deep Re-Id contributions cited in this review clearly depicting a high percentage of recent deep Re-Id works.

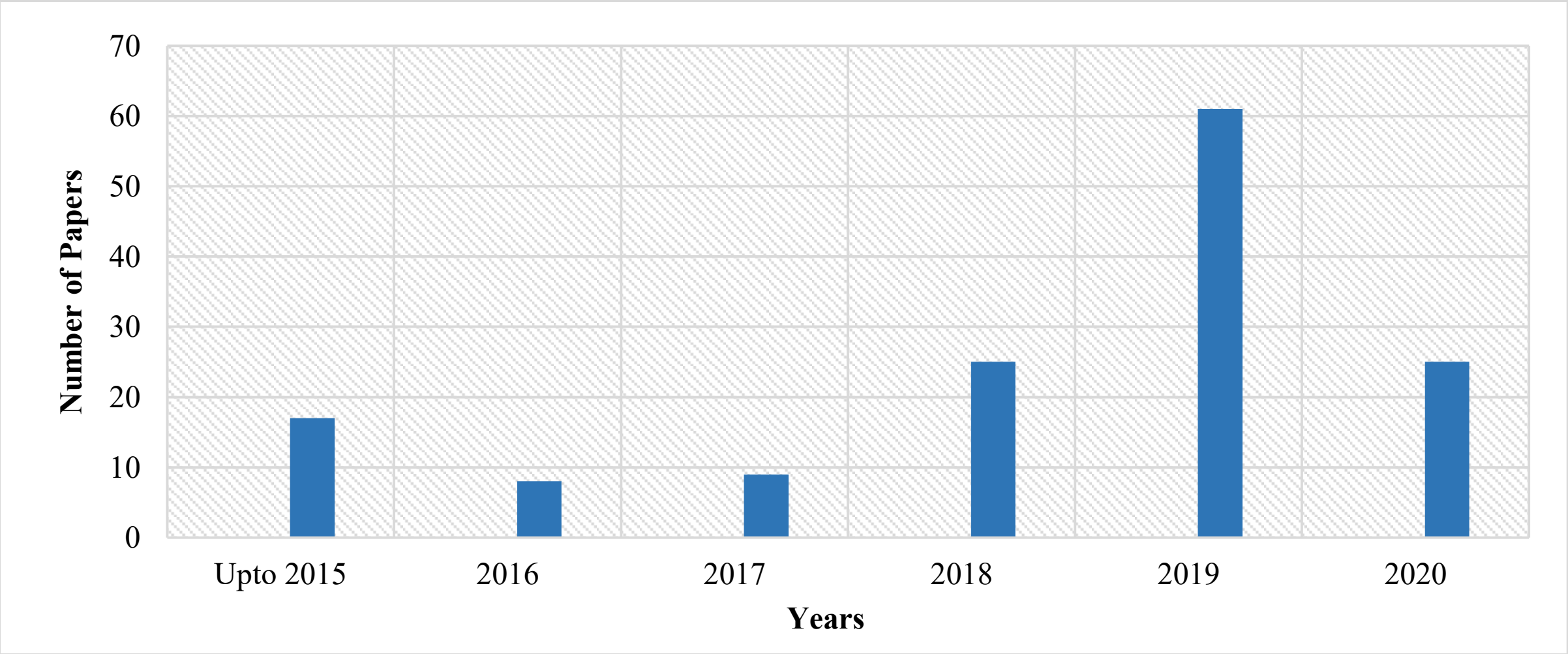

Fig. 3 Count of cited articles (year-wise)

### 2.3 Comparison with Existing Reviews

While the number of deep Re-Id implementations have grown exponentially in recent years, there are very few reviews/surveys keeping pace with the growth of deep Re-Id research. This section presents a comparative analysis of this review against some existing deep learning based Re-Id surveys. Compared to other surveys, this review includes the latest research publications (up to 2020). Table 1 presents Re-Id comparisons on the basis of several deep learning aspects such as architecture types, Re-Id datasets, challenges, modality, cross-domain approaches and metric learning methods. The green rows indicate the presence of theoretical analysis while the orange rows indicate the presence of tabular information.

TABLE 1 COMPARISON OF THIS CURRENT REVIEW WITH THE EXISTING DEEP LEARNING BASED RE-ID REVIEWS/SURVEYS. ROWS IN GREEN INDICATE THEORETICAL ANALYSIS WHILE ORANGE ROWS DEMONSTRATE TABULAR INFORMATION

| Deep Re-Id Reviews/Surveys | | | Wang et al. [4] | Almasawa et al. [5] | Wu et al. [6] | Islam [7] | Current Review |
|---|---|---|---|---|---|---|---|
| Year of Publication | | | 2018 | 2019 | 2019 | 2020 | - |
| Year of Latest Citation | | | 2018 | 2019 | 2018 | 2020 | 2020 |
| Re-Id Datasets | | Theory | | ✓ | ✓ | ✓ | ✓ |
| Re-Id Datasets | | Table | | ✓ | | ✓ | ✓ |
| Image & Video Re-Id Methods | | Theory | | ✓ | | | ✓ |
| Image & Video Re-Id Methods | | Table | | ✓ | | | ✓ |
| Deep Re-Id Analysis | Architecture Types | Theory | ✓ | | ✓ | ✓ | ✓ |
| Deep Re-Id Analysis | Architecture Types | Table | | | | ✓ | ✓ |
| Deep Re-Id Analysis | Challenges | Theory | | | ✓ | | ✓ |
| Deep Re-Id Analysis | Challenges | Table | | | | | ✓ |
| Deep Re-Id Analysis | Modality | Theory | | ✓ | ✓ | | ✓ |
| Deep Re-Id Analysis | Modality | Table | | ✓ | | | ✓ |
| Deep Re-Id Analysis | Cross Domain | Theory | | | ✓ | | ✓ |
| Deep Re-Id Analysis | Cross Domain | Table | | | | | ✓ |
| Deep Re-Id Analysis | Metric Learning | Theory | ✓ | | ✓ | ✓ | ✓ |
| Deep Re-Id Analysis | Metric Learning | Table | | | | | ✓ |

Table 1 clearly shows the comprehensive nature of this review when compared to other existing surveys, presenting both theoretical and tabular analysis of various deep Re-Id aspects.

## 3. Benchmark Datasets for Person Re-identification

Several benchmark datasets have been collected over the years to train and test the robustness of person re-identification systems. These datasets are useful in validating different re-identification approaches in terms of recognition accuracy. Table 2. gives detailed information about various re-id datasets.

TABLE 2 RE-ID BENCHMARK DATASETS

| Dataset Type | Dataset | Release Year | Camera | Identity Count | Images/Videos Count | Challenges | Top Results - Rank 1 Accuracy (%) |
|---|---|---|---|---|---|---|---|
| Image Re-Id Datasets | VIPeR [8] | 2007 | 2 | 632 | 1264 images | Viewpoint variation | 67.21 [9] |
| Image Re-Id Datasets | iLIDS [10] | 2009 | 2 | 119 | 476 images | Illumination variation and occlusion | 82.20 [11] |
| Image Re-Id Datasets | GRID [12] | 2010 | 8 | 1800 | | Viewpoint, lightning and color variations, occlusion | 28.00 [13] |
| Image Re-Id Datasets | CAVIAR4RE-ID [14] | 2011 | 2 | 72 | 720 images | Resolution, light and pose variations, occlusion | 53.60 [15] |
| Image Re-Id Datasets | CUHK01 [16] | 2012 | 2 | 971 | 1942 images | Pose, viewpoint and lightning variations | 98.73 [17] |
| Image Re-Id Datasets | RGBD-ID [18] | 2012 | 1 (RGB-D) | 79 | 316 images | View and clothing variation for same identities | 76.70 [19] |

| Dataset Type | Dataset | Release Year | Camera | Identity Count | Images/Videos Count | Challenges | Top Results - Rank 1 Accuracy (%) |
|---|---|---|---|---|---|---|---|
| | CUHK02 [20] | 2013 | 10 | 1816 | 7264 images | Illumination and pose variations, partial occlusion | - |
| | CUHK03 [21] | 2014 | 2 | 1360 | 13164 images | Alignment variation, occlusion, missing body parts | 96.43 [2] |
| | Market-1501 [22] | 2015 | 6 | 1501 | 32668 images | Illumination, scale and pose variations, partial occlusion | 95.34 [17] |
| | Kinect-Re-Id [23] | 2015 | 1 RGB-D | 71 | 483 videos | Viewpoint and lightning variations | 99.40 [19] |
| | DukeMTMC-Re-Id [24] | 2017 | 8 | 1852 | | Illumination, view and pose variations, background clutter, occlusion | 88.19 [17] |
| | RegDB [25] | 2017 | 1 RGB<br>1 IR | 412 | 4120 images<br>4120 thermal images | Pose, distance and lightning variation | 48.43 [26] |
| | SYSU-MM01 [27] | 2017 | 4 RGB<br>2 IR | 491 | 287628 images<br>15792 IR images | Color and exposure variation | 65.10 [28] |
| | Airport [29] | 2019 | 6 | 9651 | 3.13 images per person on average | Viewpoint and illumination variation, detection error, occlusion, background clutter | - |
| Video Re-Id Datasets | ETHZ [30] | 2007 | 2 | | | Partial occlusion | 93.00 [31] |
| | PRID2011 [32] | 2011 | 2 | 983 | 100 to 150 images per individual | Viewpoint, pose, lightning and background variations | 98.80 [33] |
| | 3DPES [34] | 2011 | 6 | 200 | 500 videos | | 72.23 [35] |
| | iLIDS-VID [36] | 2014 | 2 | 300 | 600 videos, 73 frames per video on average | Viewpoint and lightning variation, occlusion, cluttered background, similar clothing for different identities | 88.00 [37] |
| | MARS [38] | 2016 | 6 | 1261 | 20715 videos | Viewpoint and pose variations, partial occlusion, detection error, small inter-class and large intra-class variation | 87.30 [39] |
| | DukeMTMC-VideoRe-Id [40] | 2019 | | 1812 | 12 frames per second in a tracklet, 2196 training tracklets with 369656 frames, 2636 testing tracklets having 445764 | Lightning, pose and viewpoint variation, noisy background, occlusion | - |

Re-Id datasets can be broadly classified into two types: image-based and video based datasets.

### 3.1. Image Based Re-Id Datasets

The VIPeR [8] dataset has been created for viewpoint invariant pedestrian recognition. It contains 1264 images of 632 identities that have been captured from 2 cameras. The iLIDS [10] dataset has been acquired at an airport arrival hall and has 476 images of 19 identities from 2 cameras. The GRID [12] dataset is acquired from a busy underground train station having images of 800 identities from 8 cameras. CUHK03 [21] is another large scale re-id dataset having 13164 images of 1360 identities collected using 2 cameras.

Datasets like Market-1501 [22], DukeMTMC-Re-Id [24] and CUHK02 [20] have used 6,8 and 10 cameras respectively, thereby increasing the number of camera views used to collect people images. A few multi-modal re-id datasets also exist such as the RGBD-ID [18], KinectRe-Id [23] based on RGB-depth images and RegDB [25], SYSU-MM01 [27] containing both RGB and infrared images. The largest image based re-id dataset is the Airport [29] collected from 6 cameras of a mid-sized airport containing images for 9651 identities. Fig. 4 shows some sample images from the Market-1501 dataset.

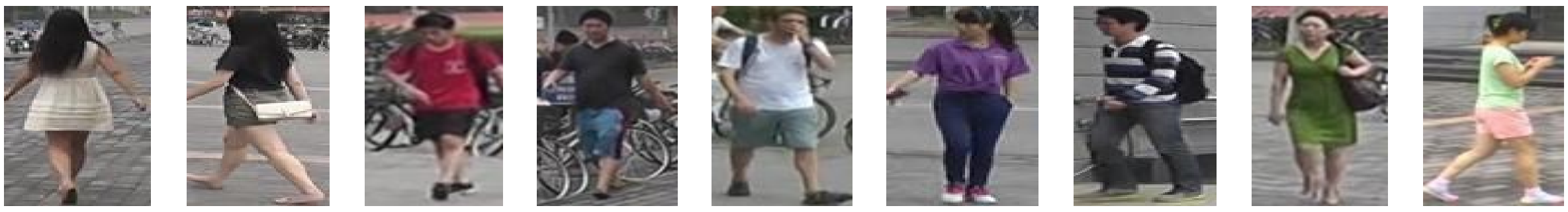

Fig. 4 Sample Images from Market-1501 dataset

### 3.2. Video Based Re-Id Datasets

PRID2011 [32] dataset contains 983 identities collected from 2 cameras. iLIDS-VID [36] contains 600 videos of 300 people from 2 cameras. Another popular video based re-id dataset is the large scale dataset MARS [38] which contains 20715 videos of 1261 identities from 6 cameras. The most recent addition to the video re-id dataset is the DukeMTMC-VideoRe-Id [40] containing 1812 identities.

Considering the growth of re-id datasets over the years, several inferences can be made. *Firstly*, Re-Id datasets have grown both in number and scale which is a great benefit since deep models require large amount of samples for effective training. *Secondly*, the variety of samples within these datasets present numerous re-id challenges such as variations in pose, lightning and scale, occlusion, background clutter, same people wearing different clothing (large intra-class disparity) or different people wearing similar clothing (small inter-class disparity), thereby allowing deep models to learn effective generalization of appearance. *Thirdly*, very few datasets are multi-modal [23], [18] leading to an over reliance on RGB image and video based approaches. *Fourthly,* in a supervised environment, deep models require labelled samples for learning. As the size of datasets grow, it becomes less feasible to annotate them manually. While most of the datasets discussed above have manually annotated samples, datasets like Market-1501 or CUHK03 have used Deformable Part Model (DPM) [41] for sample labelling.

## 4. Image Based Deep Re-Id Contributions

This section details the recent image based deep Re-Id contributions. These contributions have been categorized according to the following aspects: 1) Architecture types for Re-Id 2) Re-Id challenges 3) Data Modality for Re-Id 4) Cross-Domain Re-Id 5) Metric Learning for Re-Id. These categories are not exclusive and often overlap in various implementations but each has a distinct conceptual aspect to it.

### 4.1 Deep Re-Id Architecture Types

This section talks about the different kinds of architectures used for deep learning based re-identification. Specifically, the Re-Id contributions have been categorized as 1) Classification Models 2) Verification Models 3) Triplet Based Models 4) Part-Based Models 5) Attention Based Models as shown in Fig. 5.

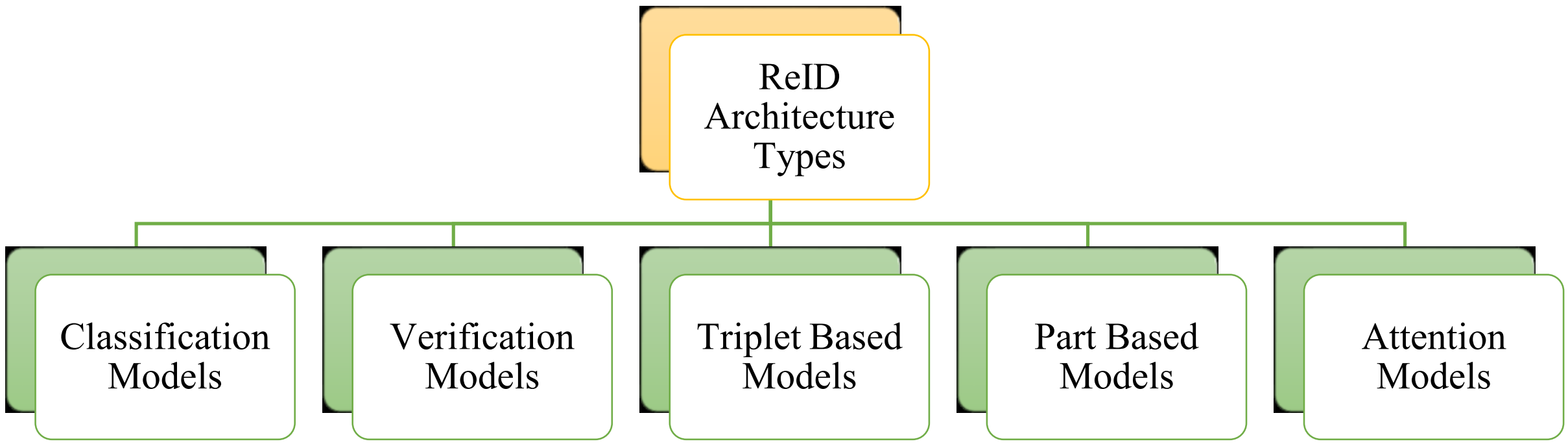

Fig. 5 Different Kinds of Deep Architectures used for Re-Id methods

### 4.1.1 Classification Models for Re-Id

Classification models (also known as Identification models) consider Re-Id as a multi-class classification problem [42]. Given a dataset with a finite number of identities and each identity having a number of samples, these models are trained using identity labels from samples. Classification models can be formally described as follows. Let there be an image gallery of $\kappa$ people $P = \{p_1, p_2, p_3 ... p_\kappa\}$ with identity labels $L = \{l_1, l_2, l_3 ... l_\kappa\}$. The model is trained with labels $L$ from various sample identities of people $P$. After training, given a query sample $p_x$ having identity $l_x$ the classification model tries to output a high score for label $x$ and low score for all other identity labels. Since Re-Id models are trained on samples from various identities, they require large number of samples per identity to capture diverse features from each individual. Lack of diverse samples often lead to over-fitting. The softmax loss is usually employed in classification models which encourages the separation of different identity samples [43]. However, Re-Id presents large intra-class variations such as pose variations, view variations, lightning variations, occlusion, background clutter etc, for which the softmax loss has performed poorly. Several improvements have been suggested over the softmax loss to handle these intra-class challenges.

Zhu et al. [43] aim to overcome the inability of softmax loss to handle intra-class variations by using it in conjunction with the center loss [44] which was originally used for facial recognition. Authors train a convolutional neural network (CNN) with the proposed combination of softmax and center loss to extract discriminative features and obtain better Re-Id results as shown in Fig. 6.

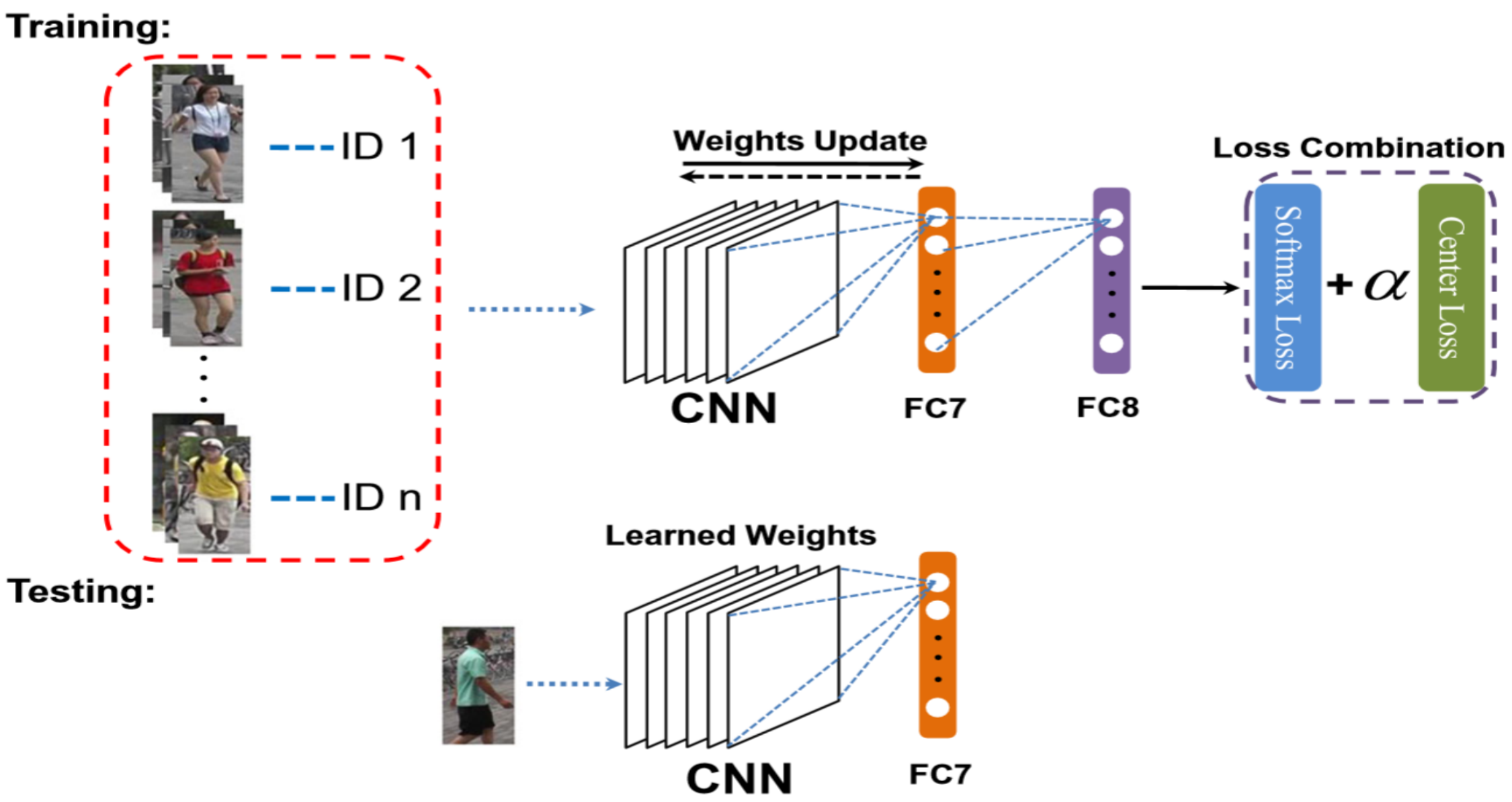

Fig. 6 CNN training based on combination of Softmax and Center loss *[43]*

Zhong et al. [42] enhance the Re-Id classification performance by using a multi-loss training setup having a combination of softmax loss, center loss and a novel "inter-center loss" . While the softmax loss differentiates between different identity samples, the center loss brings same class identities closer to their center and the inter-center loss maximizes the distance between centers of different identity as shown in Fig. 7.

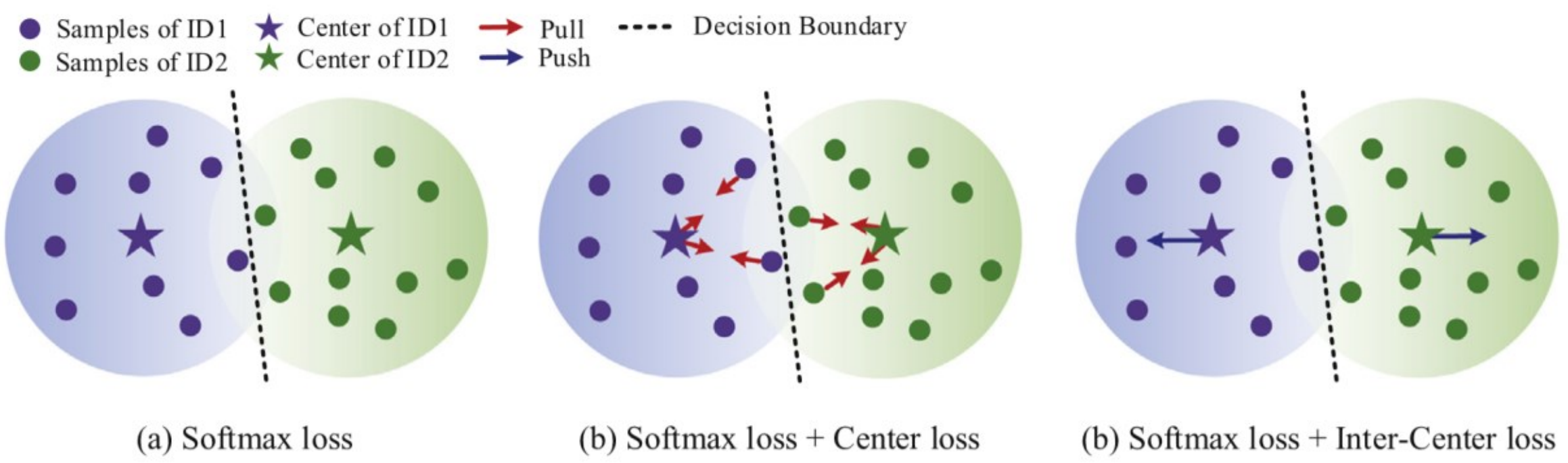

Fig. 7 a) Softmax loss separates different identity samples. b) Center loss pulls same class samples closer to their center. c) Inter-Center loss pushes different identity centers further apart *[42]*

Fan et al. [45] propose a novel "Sphere Softmax Loss" by modifying the softmax loss. Instead of mapping sample images to a Euclidean space embedding, sphere loss maps sample images to the surface of a hypersphere manifold, thereby limiting spatial distribution of data points to angular variations. The proposed loss trains using the angle between sample vector and target class vector.

### 4.1.2 Verification Models for Re-Id

Verification Models consider Re-Id to be a binary-classification problem. Given a pair of images, these models classify them to be either same or different. These models implement a pair of CNNs to extract features from input pair and compare their similarity. Verification models use the contrastive loss [46] which was first used for dimensionality reduction. In Re-Id, the contrastive loss tries to pull same identity pairs to zero distance in the feature space while pushing different identity pairs beyond a given margin. Verification models suffer from the *class imbalance problem*. Consider a dataset having K number of identities, each of which has M number of image samples. The dataset is balanced with respect to different identity samples. However, if we consider the number of positive and negative samples present with respect to a single identity, there are M positive samples and (K-1)M number of negative samples. This leads to class imbalance in training verification based models.

Zhang et al. [47] combine the verification and classification Re-Id models to learn "deep features from body and parts" (DFBP). Specifically, the verification model is implemented using two neural networks that are trained by comparing body parts from image sub-regions of input pair while the global region features are extracted from a classification model to learn body-based features. The concatenated body-based and part-based features form the final representation. Zhong et al. [48] propose a novel "Feature Aggregation Network" (FAN) which also combines the classification and verification tasks. FAN extracts multi-level CNN features from input image pairs. Then, Recurrent Comparative Network (RCN) containing attention module compares appearance of input image pairs for verification loss. The CNN features are pooled directly using the Global Average Pooling (GAP) for classification loss. Fig. 8 shows the proposed model.

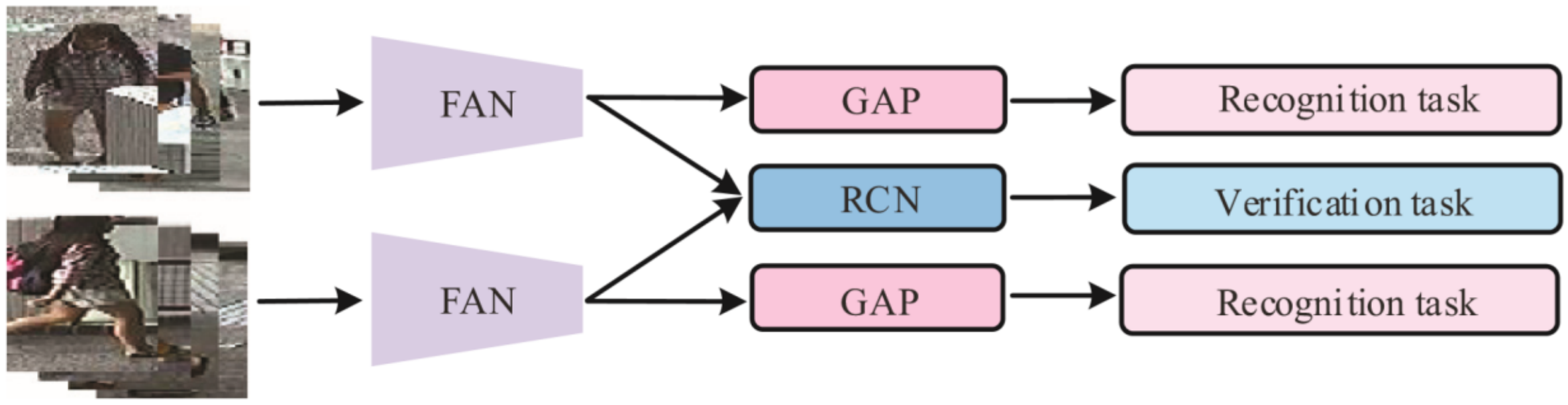

Fig. 8 Overview of proposed hybrid model. CNN features are extracted from input image pair using FAN networks. RCN learns joint representation for verification task while GAP is used for recognition task *[48]*.

### 4.1.3 Triplet Based Re-Id Models

Triplet models for Re-Id take triplet input units. Each triplet unit contains three image samples: the anchor, a positive sample (having same identity as the anchor) and a negative sample (different identity from the anchor). The triplet loss [49] is trained to keep the Euclidean distance between anchor and positive sample less than anchor and negative sample. Let $T_i$ represent the ith triplet such that $T_i = \langle T_i^a,\ T_i^p,\ T_i^n \rangle$ having anchor image $T_i^a$, positive sample $T_i^p$ and negative sample $T_i^n$. $\Gamma(I)$ represents the extracted CNN features for image I. $\|x\|$ represents the $\mathcal{L}_2$ norm. The proposed triplet loss enforces the following condition:

$$\left\|\Gamma(T_i^a) - \Gamma\left(T_i^p\right)\right\| < \|\Gamma(T_i^a) - \Gamma(T_i^n)\| \quad (1)$$

Fig. 9 demonstrates the preservation of condition (1).

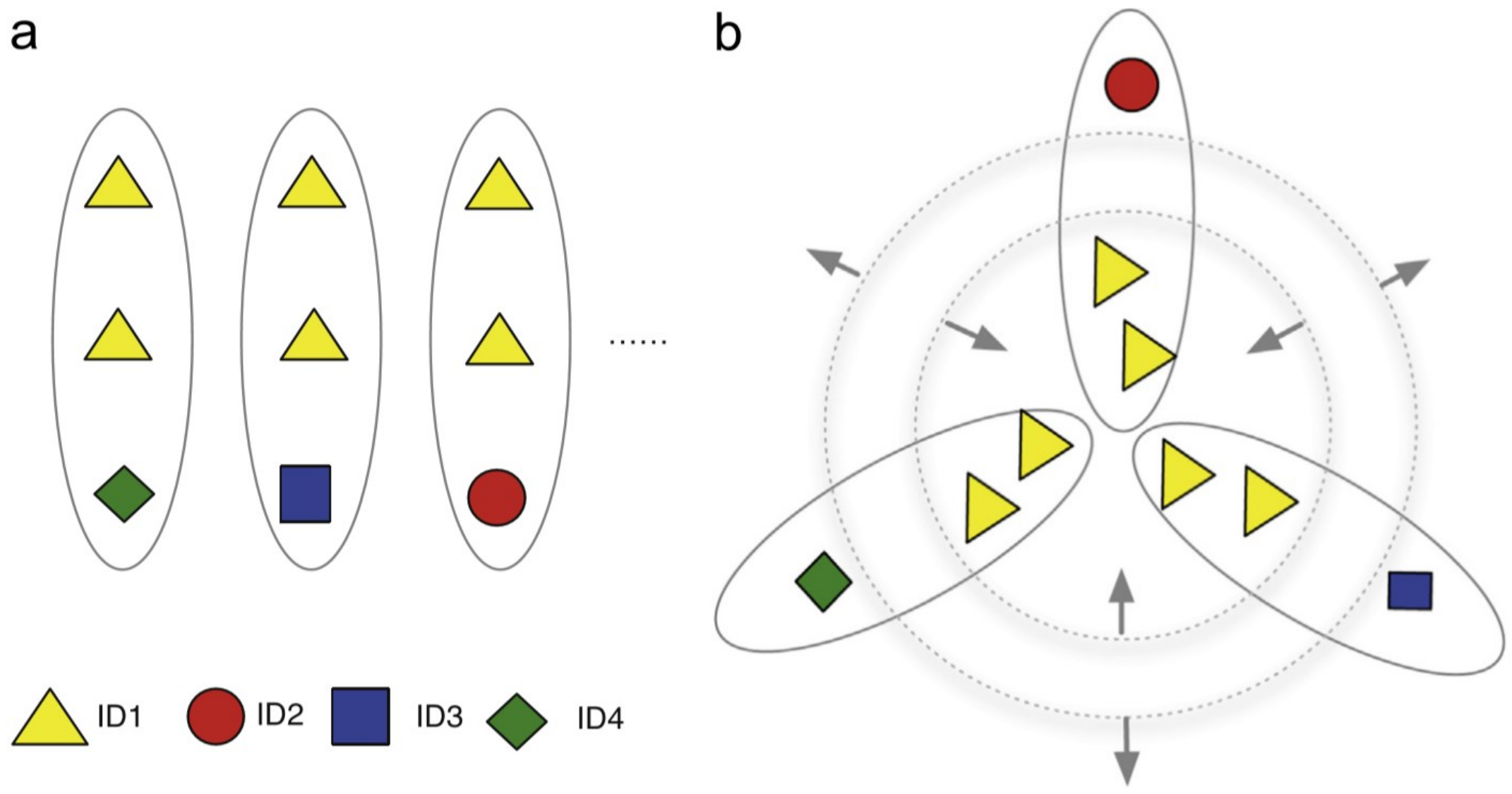

Fig. 9 a) Three triplet units having anchor identity (yellow triangle), positive sample (also yellow triangle) and negative sample (green/purple/red shapes). b) The Triplet Loss enforces the features of positive samples closer meanwhile pushing the negative samples away *[49]*

The main drawback of triplet models is that they only use weak annotations from a triplet to learn discriminative features, unlike a classification model learning from numerous samples of a given identity available in the dataset [50]. The traditional triplet loss has faced the issue of slow convergence and hence, several improvements have been formulated to improve the discrimination ability of triplet based models. Table 3 demonstrates these novel improved triplet losses.

TABLE 3 TRIPLET LOSS IMPROVEMENTS FOR RE-IDENTIFICATION

| References | Year | Triplet Loss Improvements | Benefit |
|---|---|---|---|
| Ding et al. [49] | 2015 | Triplet Loss | Works on relative distance among intra-class and inter-class identities. |
| Cheng et al. [51] | 2016 | Improved Triplet Loss | Ensures intra-class compactness |
| Hermans et al. [52] | 2017 | Batch Hard Triplet Loss | Removes triplet mining step overhead. Learning from similar inter-class and varying intra-class samples |
| Zhu et al. [53] | 2017 | Hash Based Triplet Loss | Ensures that the hamming distance of hash codes from the anchor and positive sample is less than that of anchor and negative sample |
| Su et al. [54] | 2018 | Attribute Triplet Loss | Learns a large number of human attributes considering their contextual cues. |
| Wu et al. [50] | 2019 | OIM + Improved Triplet Loss | Learn similarity metric and fully utilize label information of samples |
| Yuan et al. [55] | 2019 | Mini-Cluster Loss | Reduce intra-class and enlarge inter-class differences |
| Si et al. [56] | 2019 | Compact Triplet Loss | Reduce intra-class and enlarge inter-class differences |
| Yang et al. [57] | 2019 | Adaptive Nearest Neighbour Loss (ANN) | Solves slow convergence and local optima for triplet based loss |
| Choe et al. [31] | 2019 | Mixed Distance Maximization Loss | Maximize intra-distance and keep triplet distance larger than sample distance |
| Zhou et al. [1] | 2019 | Symmetric Triplet Loss | Gradients derived for positive samples are symmetric allowing consistent minimization of intra-class distances |
| Fan et al. [45] | 2019 | Sphere Softmax Loss | Maps image samples to a hypersphere manifold |
| Zhang et al. [58] | 2020 | Hybrid Triplet Loss (HTL) | Learns domain invariant and camera invariant properties |
| Jiang et al. [59] | 2020 | Weighted Triple-Sequence Loss (WTSL) | Reduce impact of outlier frames in video sequences |

| References | Year | Triplet Loss Improvements | Benefit |
|---|---|---|---|
| Zhang et al. [60] | 2020 | Wasserstein Triplet Loss | Uses the Wasserstein Distrance to rearrange global distance between samples |
| Sikdar et al. [61] | 2020 | Batch Adaptive Triplet Loss | Exponential learning from hard positives compared easier positives in triplet scheme |

### 4.1.4 Part-Based Re-Id Models

In the initial years, deep Re-Id methods focussed mostly on extracting global image-level features to identify individuals. However, this approach quickly became ineffective in handling small inter-class variations such as identifying different people wearing same color clothes. This has led to a gain in popularity of part-based Re-Id methods due to their superior discrimination capability based on finer part-level cues which are usually suppressed while extracting global features [2]. Part-Based Re-Id methods extract different image regions to find discriminative part-level features.

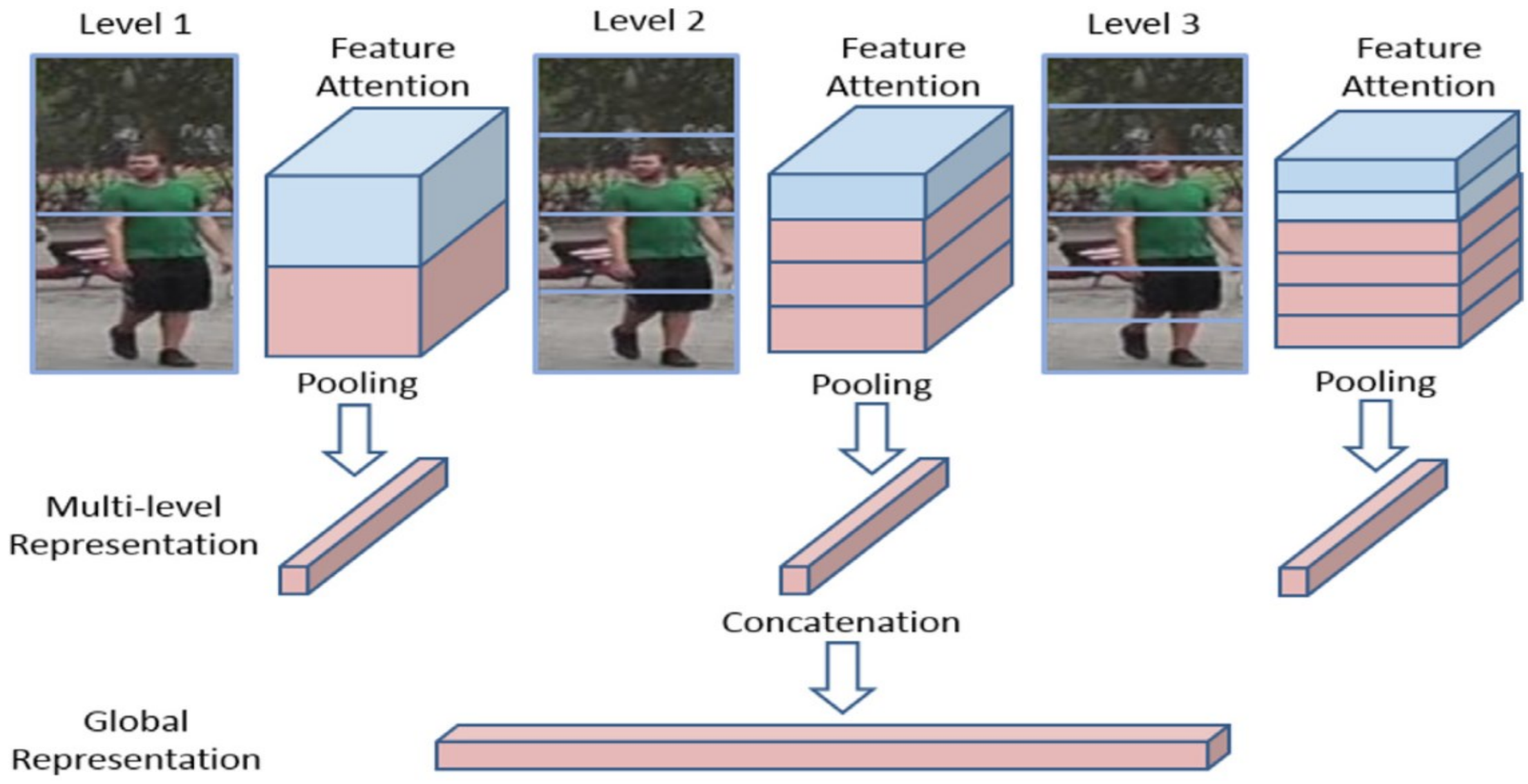

Fig. 10 Multiple feature attention networks used to generate different levels of part features. Extracted feature vectors are concatenated to obtain global representations *[62]*

Yan et al. [62] propose a feature attention block for part-based Re-Id. The authors slice features maps into spatial features and assign them weights thereby highlighting the important part regions as demonstrated in Fig. 10. Tian et al. [63] propose a novel Strong Part Based Network (SP_Net) that divides feature maps into ℕ parts, thereby learning part level features using ℕ part losses combined to obtain a local loss. The local loss is combined with global loss in a weighted manner to obtain discriminative capabilities. The major challenges faced by part-based models are the variations in pose, alignment and scale of corresponding image regions (parts) under comparison.

### 4.1.5 Attention-Based Re-Id Models

Attention based Re-Id models aim to selectively choose regions of high interest from input information. The proposed "Attention Modules" focus on extracting regions containing highly discriminative features while ignoring other regions having little or no discriminative capability. Such an approach of targeting specific regions helps to overcome Re-Id challenges like background clutter, misalignment etc [64]. Attention models have proven their superior performance in several computer vision applications including Re-Id with the growth of Recurrent Neural Networks (RNN) based on Long Short Term Memory (LSTM) [65]. Various Re-Id implementations have incorporated the attention mechanism to enhance their performance. Table 4 presents an analysis of contributions of attention modules in Re-Id methods.

TABLE 4 CONTRIBUTIONS OF ATTENTION MODULES IN RE-ID METHODS

| Reference | Year | Region of Attention | Benefit |
|---|---|---|---|
| Yang et al. [2] | 2019 | Whole body and body parts | Discriminative feature extraction |
| Bao et al. [66] | 2019 | Body parts | Robust to part misalignment and background clutter |
| Zhou et al. [1] | 2019 | Image foreground | Robust to background clutter |
| Wu et al. [67] | 2019 | Spatial regions in video frames, temporal pooling over entire video | Extract discriminative Re-Id features from essential frames within video |
| Li et al. [68] | 2019 | Convolutional features | Learns interdependence of channels within convolutional features |
| Zhang et al. [37] | 2019 | Video frames | Select informative frames for dimensionality reduction of features |
| Wan et al. [69] | 2019 | Spatial region of input images | Local parts discovery |
| Tay et al. [70] | 2019 | Physical appearance attributes like gender, hair, upper clothing color, lower clothing color, pant etc | Discriminative representation of identity based on appearance attributes (Fig. 11) |
| Hou et al. [71] | 2019 | Generated Frames created by GAN generator | Generate occluded regions in video frames using temporal attention module to remove occlusion |
| Subramaniam et al. [72] | 2019 | Common visual features across frames of a video | Robust to background noise and extracts common features in video frames |
| Chen et al. [73] | 2019 | Spatial and Channel-wise attention | Improved attention quality due to self-critical attention learning |
| Zhang et al. [74] | 2020 | Video frames focussed at multiple scales | Spatio-temporal video representations |
| Li et al. [75] | 2020 | Global image features at multiple scales | Robust to spatial misalignment and local feature dependencies |
| Qian et al. [17] | 2020 | Spatial regions of input, multi-scale features | Discriminative feature extraction |
| Zhang et al. [60] | 2020 | Global and local features | Robust to misalignment, helps to distinguish between important and misleading parts |

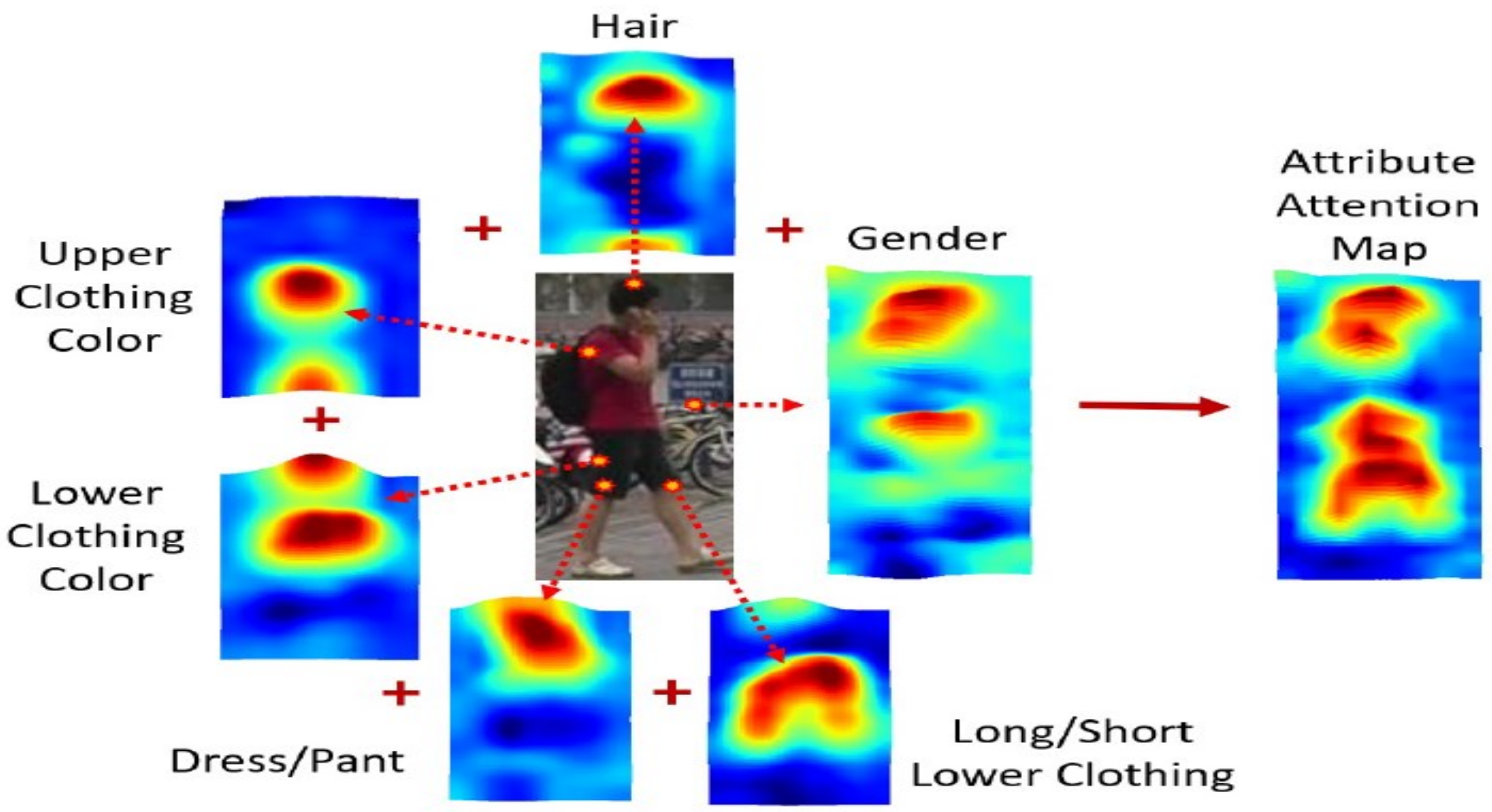

Fig. 11 Attribute Attention Map (AAM) generated from six heat maps corresponding to attributes such as gender, hair, clothing etc *[70]*

Table 5 gives a detailed overview of various deep Re-Id contributions based on their architecture types.

TABLE 5 RE-ID CONTRIBUTIONS BASED ON DIFFERENT ARCHITECTURES

| Reference | Year | Architecture | Key idea | Loss Function |
|---|---|---|---|---|
| Ding et al. [49] | 2015 | Triplet | Propose "Triplet loss" the utilizes three samples to learn discriminative features and also propose a triplet generation scheme | Triplet loss |
| Zhu et al. [43] | 2017 | Classification | Combine Softmax and Center Loss [44] to improve discriminative capability of CNN features | Softmax + Center Loss |
| Zhu et al. [53] | 2017 | Triplet + Part | Propose a "Part Based Deep Hashing" (PDH) network that integrates hashing for higher efficiency of large scale Re-Id. Implement a triplet loss that reduces the Hamming distance of positive sample parts and increase that of negative sample parts. | Triplet loss |
| Huang et al. [76] | 2017 | Part | Propose a novel method DeepDiff to evaluate similarity between corresponding parts using original data, feature maps and spatial variations from three deep subnets | Softmax loss |
| Zhu et al. [43] | 2018 | Classification | Combine softmax and center loss to overcome intra-class variations in samples | Softmax loss + Center loss |
| Koo et al. [77] | 2018 | Part | Use information from face and body to obtain discriminative representations in indoor camera surveillance environment | Softmax loss |
| Tao et al. [78] | 2018 | Triplet | Propose a "Deep Multi-View Feature Learning" (DMVFL) method that fuses handcrafted and deeply learned features to obtain robust representations | Triplet loss |
| Su et al. [54] | 2018 | Triplet | Propose a three-stage "Weakly Supervised Multi-Type Attribute Learning Framework" using a novel "Attribute Triplet Loss" to predict visually invariant features containing contextual cues | Attribute triplet loss |
| Zhang et al. [47] | 2018 | Classification + Verification + Part | Learn deep features from body parts and entire body using verification and classification models respectively to obtain final representation | Softmax loss |
| Zhong et al. [42] | 2019 | Classification | Propose a novel "Inter-Center Loss" to improve the discriminative capability of CNN features | Softmax + Center + Inter-Center Loss |
| Fan et al. [45] | 2019 | Classification | Propose a novel CNN based network "SphereRe-Id" adopting a novel sphere loss mapping sample images to hypersphere manifold and a balanced sampling strategy to address class imbalance | Sphere Softmax Loss |
| Bao et al. [66] | 2019 | Classification + Attention | Propose a dual-branch CNN based network having a global branch to process features from overall human body and an attention branch to selectively focus on attentional part of information from input | Softmax Loss |
| Yang et al. [2] | 2019 | Part based + Attention | Introduce a novel attention driven multi-branch network to learn discriminative representations from whole body and body parts focussing on spatial and channel-wise information | Softmax loss |
| Bao et al. [66] | 2019 | Classification + Attention | Combine global features with attention focussed discriminative features to reduce impact of misalignment and background clutter | Softmax loss |
| Yan et al. [62] | 2019 | Part + Attention | Propose a feature attention block for Re-Id that pays attention to sliced part features in a weighted manner to highlight the most discriminative part regions which are robust to misalignment | Softmax loss |
| Zhou et al. [1] | 2019 | Triplet + Part + Attention | Propose a novel "Foreground Attentive Neural Network" (FANN) that utilizes a foreground attentive subnetwork and a novel regression loss function to learn foreground regions which is then fed to body part features using novel symmetric triplet loss | Regression loss + Symmetric Triplet loss |
| Wu et al. [3] | 2019 | Classification + Triplet | Combine classification loss, triplet loss and center loss to constrain Euclidean distance of same identity samples closer than those of different identities | Softmax loss + Triplet loss + Center loss |
| Zhang et al. [79] | 2019 | Classification + Triplet | Introduce a dual-branch "Multi Branch Slice-Based Network" (MSN) learning multi-level local and global features using a novel "triplet-center loss" | Triplet-Center loss |
| Zhao et al. [80] | 2019 | Triplet | Introduce a multi-level triplet model MT-net, extracting multi-level features which are both global and local | Triplet loss |
| Yuan et al. [81] | 2019 | Classification + Triplet | Introduce a deep joint embedding learning framework that uses classification and an improved triplet loss. The improved triplet loss works on hard triplets generated | Softmax loss + Improved Triplet loss |

| Reference | Year | Architecture | Key idea | Loss Function |
|---|---|---|---|---|
| Wu et al. [50] | 2019 | Classification + Triplet | Combine classification and triplet loss to make full use of labels as well as learn similarity measure simultaneously | Online Instance Matching loss + Triplet loss |
| Tian et al. [63] | 2019 | Part | Propose a two-branch CNN to combine learning global and local part level features simultaneously | Softmax loss |
| Li et al. [68] | 2019 | Classification + Verification + Attention | Propose a novel network with attention module to highlight essential features and a multi-loss function to reduce intra-class distance and increase inter-class distance | Cross-entropy loss + novel verification loss |
| Yuan et al. [55] | 2019 | Triplet | Propose a novel "mini-cluster" loss that ensures the largest distance of same identity samples (inner divergence) to be less than the smallest distance of different identity samples (outer divergence) | Mini-cluster loss |
| Fan et al. [45] | 2019 | Classification | Propose a novel classification CNN called SphereRe-Id that uses a novel "Sphere Softmax loss" mapping samples to a hypersphere manifold | Sphere Softmax loss |
| Ling et al. [82] | 2019 | Classification + Verification | Introduce MTNet with four losses for identification and verification of person identity and person attributes | Softmax loss |
| Si et al. [56] | 2019 | Classification + Triplet | Propose a novel "Compact Triplet Loss" that improves the batch hard triplet loss to reduce intra-class variations and increase inter-class differences. Combine with a classification loss for better discrimination capability | Compact triplet loss + Softmax loss |
| Tian et al. [83] | 2019 | Verification + Triplet | Propose a novel "Adaptive Verification Loss" (ADV loss) that learns only from meaningful hard sample pairs mined by using a weighted triplet loss. | ADV loss + Triplet loss |
| Zhong et al. [48] | 2019 | Classification + Verification | Introduce a novel "Feature Aggregation network" (FAN) network to learn features from various layers of deep network along with multiple losses | Softmax loss |
| Wang et al. [84] | 2019 | Part | Propose a novel "Part-Based Pyramid Loss" that takes quadruplet input samples and learns body part features using relationship of distance and angle among samples | Part based pyramid loss |
| Yao et al. [85] | 2019 | Classification + Part | Propose a "Part Loss Network" (PL-net) that trains on body part and global features | Classification loss + Part loss |
| He et al. [86] | 2019 | Classification + Verification | Adopt the "lifted structured loss" due to its superiority to contrastive and triplet losses. Combine it with identification loss to learn relative identity information from pairs and true identity information | Softmax loss + Lifted structured loss |
| Choe et al. [31] | 2019 | Triplet | Consider intra-distance between positive samples of a triplet and distance between triplets using a novel "mixed distance" function to improve Re-Id performance | Mixed distance loss |
| Quispe et al. [87] | 2019 | Classification + Triplet | Propose a novel "Saliency Semantic Parsing Re-Id" (SSP-Re-Id) network. The Saliency guided subnetwork learns from essential parts of image while semantic parsing guided subnetwork deals with Re-Id challenges | Softmax loss + Triplet loss |
| Wan et al. [69] | 2019 | Part + Attention | Propose a novel "Concentrated SPR network" (CSPR-Net) having a "constrained attention module" to find discriminative local parts that work better than body parts and a novel "statistical-positional-relational (SPR) descriptor that gives better the performance than global features | Classification loss |
| Zhang et al. [60] | 2020 | Classification + Triplet + Attention | Propose a novel triplet loss based on the "Wasserstein distance" (Earth Mover distance) to handle the part misalignment problem using part probabilities obtained from attention maps and part features. | Softmax loss + Wasserstein triplet loss |
| Li et al. [88] | 2020 | Part | Propose a novel "Attributes-Aided Part Detection and Refinement Network" (APDR) that uses attribute learning for part localization handling the misalignment problem. Attribute features are fused to obtain discriminative features | Softmax loss + Triplet loss |
| Qian et al. [17] | 2020 | Classification + Triplet + Attention | Propose a novel "Multi-Scale Deep Architechture" (MuDeep) having a "multi-scale deep learning layer" to learn features at different scales and a "leader-based attention learning layer" to determine optimal weighting for features from each scale | Softmax + Triplet loss |
| Bai et al. [89] | 2020 | Classification + Triplet + Part | Propose a three-branch "Deep Person" framework that learns contextual body part information using LSTM and learn discriminative features using identification (global and part level) and triplet losses. | Softmax loss + Triplet loss |
| Zhang et al. [90] | 2020 | Part | Propose a dual-branch "Heterogeneous Part-Based Deep Network" (HPDN) to learn part based features using batch hard triplet and cross entropy loss | Batch hard triplet loss + Cross-entropy loss |

| Reference | Year | Architecture | Key idea | Loss Function |
|---|---|---|---|---|
| Li et al. [75] | 2020 | Classification + Triplet + Attention | Introduce a self-attention guided model that learns weighted features from different regions of human image | Softmax loss + Triplet loss |

The **highlights of deep Re-Id architectures** are:

- Deep learning based Re-Id architectures can be classification models, verification models, triplet-based models, part-based models and attention based models.
- Classification models treat Re-Id as a multi-class classification problem that use the softmax loss to predict the class of an input query. Softmax loss encourages the separation of different classes but struggles with large intra-class variations. Several Re-Id works combine classification with other model types to overcome this limitation [44], [91], [82].
- Verification models treat Re-Id as a binary classification problem, taking a pair of inputs and categorizing them as same or different. These models suffer from the class imbalance problem since the number of positive pair combinations is far less than the negative pairs where each identity contains same number of samples in the dataset.
- Triplet models input triplets of images containing an anchor, a positive and a negative sample. These models are trained on the triplet loss that aims to pull the positive sample closer and pushes the negative sample away in feature space. Several improvements have been suggested over the traditional triplet loss having convergence issues. These improvements include solving slow convergence [57], removing triplet mining overhead [52], maintaining intra-class compactness [51] etc.
- Part-based Re-Id models aim to focus on sub-regions within input to extract finer feature representations crucial in differentiating samples with small inter-class variations, usually missed in global image level representations. Focus on different parts of feature maps [62], [63], face and body regions [77], attribute guided body parts [88] etc, have yielded discriminative feature representations.
- Attention models highlight regions of high interest within input holding highly discriminative information. Gaining popularity since the growth on RNN and LSTM, attention modules have helped in implementing *spatial attention* within image/frame regions such as human body parts [66], foreground [1], physical attributes. [70] and *channel attention* within deep features [68].

### 4.2 Methods Based on Re-Id Challenges

The task of person re-identification has faced several challenges like sample variations in view, pose, lightning and scale, partial or complete occlusion, background clutter etc. Fig. 12 demonstrates some of these Re-Id challenges presented by samples from various Re-Id datasets as detailed in Table 2. Any re-id system striving to achieve competent recognition rates must be able to counter these challenges effectively. Numerous research works have been conducted with this motivation.

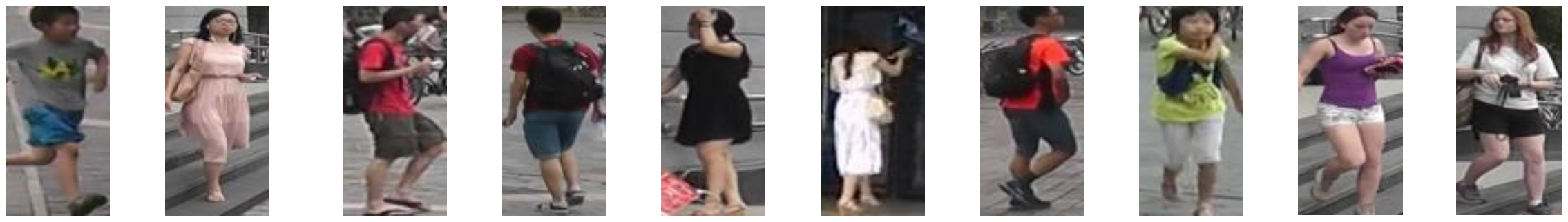

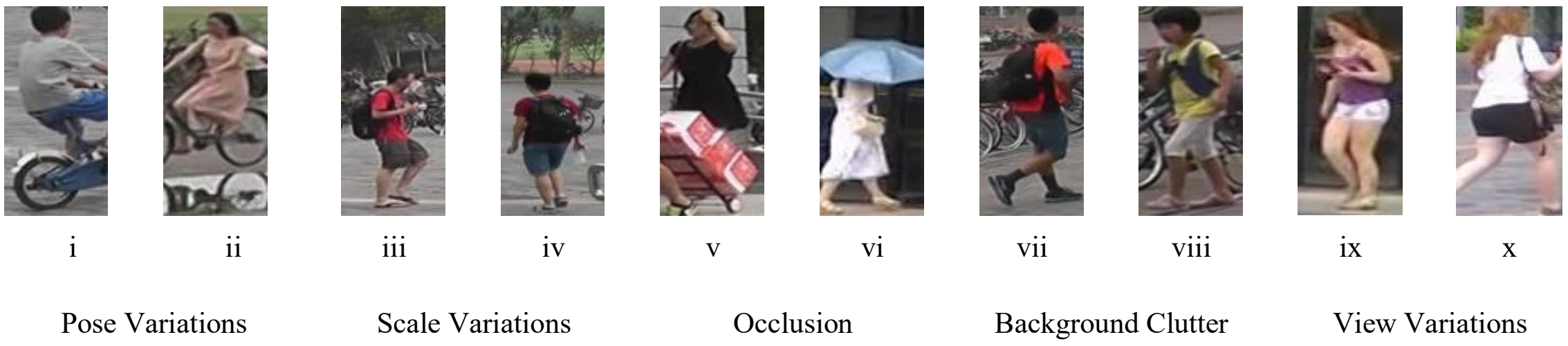

Fig. 12 . Re-Id challenges posed by dataset samples. Each column demonstrates samples from a unique identity. Columns (i) and (ii) show variations in pose, (iii) and (iv) show scale variations, (v) and (vi) display occlusion examples, (vii) and (viii) describe background clutter, (ix) and (x) show view variations in samples. All samples have been derived from the Market-1501 dataset.

Feng et al. [28] attempt to overcome the challenge of large intra-class disparity caused by view variations from images captured by cameras placed at different viewpoints. Authors propose a framework capable of learning view-specific features consistent with each camera view which utilizes a cross-view Euclidean constraint (CV-EC) and cross-view center loss to decrease the distance of features of same person from different views. Qi et al. [92] tackle lightning and viewpoint variation at two levels. Firstly, they extract deep features trained on multiple datasets which are robust to differences in illumination and view. Secondly, they use the learned features to find an optimal ensemble of metrics including the Cosine distance metric that reduces the intra-class disparity even further. Sikdar et al. [61] achieve scale-invariance by modifying the convolution functionality within a deep network. Instead of learning a kernel on a fixed scale input, the input is first transformed to a pyramid of multiple resolutions. The network then learns multiple scaled feature maps which are then re-scaled to original size before applying max pooling. Such an operation has proven to produce scale invariant results for re-id system. Input image misalignment can seriously hamper the feature learning and matching process. To handle the misalignment problem, Zheng et al. [93] introduce *pose invariant embedding* (PIE) which aligns identities within sample images to a standard pose using pose estimation. The transformed standard pose promotes learning of discriminative feature extraction and matching and is alignment invariant. Table 6 presents some novel Re-Id contributions that are robust to Re-Id challenges.

TABLE 6 CONTRIBUTIONS ROBUST TO RE-ID CHALLENGES: VARIATIONS IN POSE (P), SCALE (S), LIGHTNING (L) AND VIEW (V), OCCLUSION (O), BACKGROUND CLUTTER (B) AND MISALIGNMENT (M)

| Reference | Year | Key Idea to Avert Re-Id Challenges | Robust to Challenges | | | | | | |
|---|---|---|---|---|---|---|---|---|---|
| | | | P | S | L | O | B | V | M |
| Yu et al. [94] | 2018 | Use skeleton joint information and cloth-color type features to achieve pose and lightning invariance | ✓ | | ✓ | | | | |
| Feng et al. [28] | 2018 | Use Cross-View Euclidean constraint (CV-EC) to reduce distance of deep features of an identity from multiple views | | | | | | ✓ | |
| Zhou et al. [35] | 2018 | Propose a "self-paced learning" (SPL) method to isolate noisy samples in weighted way using model age and iteration and train CNN model with faithful samples gradually moving from easy to hard samples | | | | ✓ | ✓ | | |
| Chen et al. [95] | 2018 | Fuse attribute features learned from part-specific CNN and fuse them with Local Maximal Occurance (LOMO) features to obtain robust features | ✓ | | ✓ | | ✓ | ✓ | |
| Fu et al. [96] | 2019 | Introduce a two stream spatial segmentation network that derives spatial and fine local features | ✓ | | ✓ | | | ✓ | |
| Zheng et al. [97] | 2019 | Use a novel "alignment branch" to learn the affine transformation of high level convolutional features utilizing a spatial transformer network that crops images with too much background and pads zero borders to missing part features thereby solving scale variation and misalignment | | ✓ | | | ✓ | | ✓ |
| Chen et al. [98] | 2019 | Study correlation among cross-view visual data from multiple camera views to compose view-specific representations | | | | | | ✓ | |

| Reference | Year | Key Idea to Avert Re-Id Challenges | Robust to Challenges | | | | | | |
|---|---|---|---|---|---|---|---|---|---|
| | | | P | S | L | O | B | V | M |
| Luo et al. [99] | 2019 | Propose a novel "Dynamically Matching Local Information" (DMLI) method that automatically aligns horizontal stripes from input samples without any labelling supervision or pose estimation | ✓ | | | | | | |
| Zhou et al. [1] | 2019 | Use a novel "foreground attentive subnetwork" having a decoder trained on a novel "local regression loss" to create a binary mask suppressing background | | | | | ✓ | | |
| Qi et al. [92] | 2019 | Train CNN on six Re-Id datasets to learn robust features | | | ✓ | | | ✓ | |
| Yang et al. [57] | 2019 | Learn spatial dependencies among local regions of pedestrians in both horizontal and vertical directions using LSTM to overcome occlusion | | | | ✓ | | | |
| Zheng et al. [93] | 2019 | Propose a novel "pose invariant embedding" (PIE) by constructing a novel PoseBox through pose estimation and then training using two-stream PoseBox Fusion network | ✓ | | | | | | |
| Wei et al. [100] | 2019 | Estimate four human key points that are robust to pose and view variations. Head, upper body and lower body features are generated using these key points and a four-stream CNN is trained to generate the novel "Global Local Alignment Descriptor" (GLAD) features from both global and local regions | ✓ | | | | | | ✓ |
| Wu et al. [101] | 2020 | Use adversarial learning to learn asymmetric transformations to transform view-specific distribution to a generic view invariant feature space | | | | | | ✓ | |
| Li et al. [88] | 2020 | Use attribute learning to detect body parts thereby solving part misalignment problem | | | | | | | ✓ |
| Tang et al. [102] | 2020 | Propose a novel "Gradual Background Suppression" method that extract CNN features based on different weights assigned to body parts and background thereby suppressing the background | | | | | ✓ | | |
| Sikdar et al. [61] | 2020 | Resize input to different scales and convolve with a fixed size filter to obtain multi-resolution pyramid which is re-scaled back to fixed size to obtain scale invariant features | | ✓ | | | | | |
| Li et al. [75] | 2020 | Propose a "multi-scale attention" model evaluating important person regions in a weighted fashion and train on features fused globally and locally using cross-entropy and triplet loss | ✓ | | | ✓ | | ✓ | |

The **highlights of contributions based on Re-Id challenges** are:

- Finding a person of interest is challenging due to visual variations in pose, view, lightning, scale, partial or complete occlusion, background clutter and misalignment.
- Several deep Re-Id contributions have aimed to develop robust methodologies against these Re-Id challenges.
- Skeleton joints data and clothe colors have produced pose and lightning invariance [94], learning view-specific representations for view invariance [98], utilizing foreground attentive network to suppress noisy background [1], convolving with multi-scale input to obtain scale invariant features [61] and using pose estimation to achieve pose invariance [93] are some of efforts to overcome these challenges.

### 4.3 Re-Id Methods Based On Modality

Visible images have proven to be the most common source of discriminative information crucial for identifying individuals. Hence, research literature is filled with visible image based Re-Id methods due to their superior identification power. Despite their popularity, the visible image based methods are prone to several challenges already discussed in Section 4.2. Hence, some cross-modal Re-Id methods have also been proposed to enhance the ability of Re-Id systems further. This section discusses various visible light based and cross-modality based Re-Id methods.

### 4.3.1 Visible Image Based Re-Id Methods

Numerous visible image based Re-Id contributions have focussed upon extracting discriminative deep features to achieve high recognition rates for re-identification [103], [104]. Table 7 details some novel Re-Id image based contributions.

TABLE 7 IMAGE BASED RE-ID CONTRIBUTIONS

| Reference | Year | Key Idea | Benefit | Dataset | Rank 1 Result (%) |
|---|---|---|---|---|---|
| Wu et al. [105] | 2018 | Introduced a deep embedding approach using optimized robust features, positive mining and local adaptive similarity learning | Discriminative features | VIPeR | 49.04 |
| | | | | CUHK01 | 71.60 |
| | | | | CUHK03 | 73.02 |
| | | | | Market-1501 | 84.14 |
| Wu et al. [106] | 2018 | Introduce a multiplicative integration gating function combined with Hamadard product | Cross-view feature alignment | VIPeR | 49.11 |
| | | | | CUHK03 | 73.23 |
| | | | | Market-1501 | 67.15 |
| | | | | Market-1501 | 93.91 |
| | | | | DukeMTMC | 83.35 |
| Ke et al. [107] | 2018 | Introduce ID-AdaptNet to adapt “seen” identity features to “unseen” identities | Discriminative features | CUHK03 | 30.36 |
| | | | | Market-1501 | 81.59 |
| | | | | DukeMTMC | 67.77 |
| Zhang et al. [108] | 2019 | Use group symmetry theory to extract and utilize information from middle layers of Resnet50 (ResGroupNet) | Discriminative features | CUHK03 | 71.20 |
| | | | | Market-1501 | 92.80 |
| | | | | DukeMTMC | 86.20 |
| Jiang et al. [109] | 2019 | Introduce PH-GCN network to learn spatial relation among body parts using hierarchical graphs | Context aware discriminative features | CUHK03 | 64.90 |
| | | | | Market-1501 | 93.50 |
| | | | | DukeMTMC-Re-Id | 85.00 |
| Wu et al. [110] | 2019 | Introduce a five-branch deep model that learns body features in horizontal and vertical directions, relationship between feature channels and part features | Discriminative features | CUHK03 | 95.00 |
| | | | | Market-1501 | 94.70 |
| | | | | DukeMTMC-Re-Id | 86.70 |
| Liu et al. [111] | 2019 | Fuse Gaussian features with deep features | Fused discriminative features | Market-1501 | 84.40 |
| | | | | VIPeR | 57.20 |
| Zhang et al. [112] | 2019 | Introduce a PAAN network using layered partition strategy to fully utilize part-level and global attributes | Fused discriminative features | Market-1501 | 91.86 |
| | | | | DukeMTMC-Re-Id | 81.73 |
| Tian et al. [63] | 2019 | Introduce a two-branch RJLN network to jointly learn global and local features | Fused discriminative features | CUHK03 | 66.60 |
| | | | | Market-1501 | 93.70 |
| | | | | DukeMTMC-Re-Id | 85.50 |
| Wu et al. [113] | 2019 | Introduce a multi-branch MFML network to represent features from multiple layers | Weighted multi-layered fused features | CUHK03 | 94.40 |
| | | | | Market-1501 | 92.50 |
| | | | | DukeMTMC-Re-Id | 84.00 |
| Zhao et al. [80] | 2019 | Introduce a multi-level triplet model MT-net, extracting multi-level features which are both global and local | Fused discriminative features | CUHK03 | 79.34 |
| | | | | Market-1501 | 81.95 |
| Wang et al. [114] | 2020 | Propose a novel exclusively regularized softmax objective function | Multi-scale multi-patch features | CUHK03 | 70.40 |
| | | | | Market-1501 | 93.70 |
| | | | | DukeMTMC | 84.40 |
| Wang et al. [115] | 2020 | Fuse handcrafted features from local and global regions with deep features | Fused discriminative features | VIPeR | 52.22 |
| | | | | CUHK01 | 71.91 |

Wu et al. [110] propose a five-branch deep model that is capable of learning features not only from the usual horizontal direction but also in the vertical direction. The model scans for spatial information of body parts from left to right and head to foot, thereby learning discriminative information. Working with one type of features is often limiting in finding discriminative capabilities for re-id systems. Hence, an obvious direction of improvement is to fuse different features together to obtain higher differentiation in re-id. Zhao et al. [80] introduce a novel deep triplet model (MT-net) performing multi-level feature extraction. Both detailed and global features from each layer are combined together in an optimal proportion through training. The fused features prove to have high discriminative capabilities. While most deep learning based re-id approaches extract features only from the top layer, middle layer features can also contribute to discriminative capabilities of the model in certain situations. Wu et al. [113] introduce a *multi-level feature network with multiple losses* (MFML) which is a multi-branch architecture representing multiple middle layer representations trained on triplet loss and top layer representation trained on the hybrid loss. The representation from various layers are fused in a weighted manner based on their importance in obtaining differentiating characteristics. Based on the idea that color features hold key information in reference to re-id task, Liu et al. [111] fuse traditional Gaussian of Gaussian (GOG) features from four color channels (RGB, Lab, HSV, RnG) with deep features to obtain highly discriminative features which achieve state-of-the-art performances.

### 4.3.2 Cross-Modality Re-Id Methods:

While image based methods have proven to be most popular in the Re-Id research community, these visual feature based methods have several limitations. Visual Re-Id methods face several challenges such as variations in pose, view, lightning, scale, partial or complete occlusion, background clutter which are already discussed in Section 4.2. Also, visual Re-Id methods are highly ineffective in dark environments such as night-time, due to the poor illumination that suppresses most of the visual cues [26]. This has led to the development of multi-modal Re-Id methods that combine data from multiple modalities to reduce dependency on visual information. Table 8 lists the novel multi-modal Re-Id methods combining RGB, depth, text and infrared (thermal) based data modalities.

TABLE 8 CROSS MODALITY RE-ID CONTRIBUTIONS

| Reference | Year | Modality | Key Idea | Device |
|---|---|---|---|---|
| Ren et al. [116] | 2017 | RGB + Depth | Fuse anthropometric features from depth images and visual features from RGB images using a novel "multi-modal fusion layer" to obtain discriminative features | RGB Camera + Kinect V1 sensor |
| Feng et al. [26] | 2019 | RGB + Infrared Data | Extract "modality specific representations" (MSR) from modality-specific networks trained on a modality-specific loss to learn discriminative features from each domain. Use a cross-modality Euclidean constraint to learn modality-invariant features | RGB + Infrared Cameras |
| Xiang et al. [117] | 2019 | RGB + Infrared | Propose a dual-branch neural network that fuses modality specific information from two branches using multiple granularity network (MGN) to obtain modality shared features | RGB + Infrared Cameras |
| Ren et al. [19] | 2019 | RGB + Depth | Propose a novel "uniform and vibrational deep learning" (UVDL) method that uses an auto-encoder to visible and deep features extracted from two neural nets into a common features space | RGB-D camera (Kinect) |
| Chang et al. [118] | 2020 | RGB + Textual | Combine visual CNN features with textual CNN features from textual descriptions including details of gender, clothes color and bag type information to obtain a generalized feature embedding | RGB Camera |
| Gohar et al. [119] | 2020 | Gait Data | Propose a novel "non-visual gait based" Re-Id method that uses gait data captured from accelerometer and gyroscope to learn discriminative features using Gated Recurrent Unit (GRU) to capture discriminative temporal information from input sequences | Accelerometer, Gyroscope |
| Wang et al. [120] | 2020 | RGB + Infrared | Propose a novel "multi-patch networking network" (MPMN) that utilizes a single deep neural net to process both RGB and thermal images. A novel "multi-patch modality alignment" loss mines for hard subspaces. A novel "cross-patch correlation distillation" (CPCD) loss enforces cross-patch similarity to boost cross-modality embedding. A novel "patch aware priority attention" (PAPA) method prioritizes training of hard patch tasks over others. | RGB + Infrared Cameras |

Chang et al. [118] learn a similarity metric for visual and textual representations as demonstrated in Fig. 13. Authors use the Resnet architecture to extract visual features from image samples. A 2500-dimensional textual feature embedding is extracted from textual description of each sample (gender, clothe color etc) using tokenization, lemmatization and stemming. The model is trained in an end to end manner using a triplet architecture on both visual and textual representations.

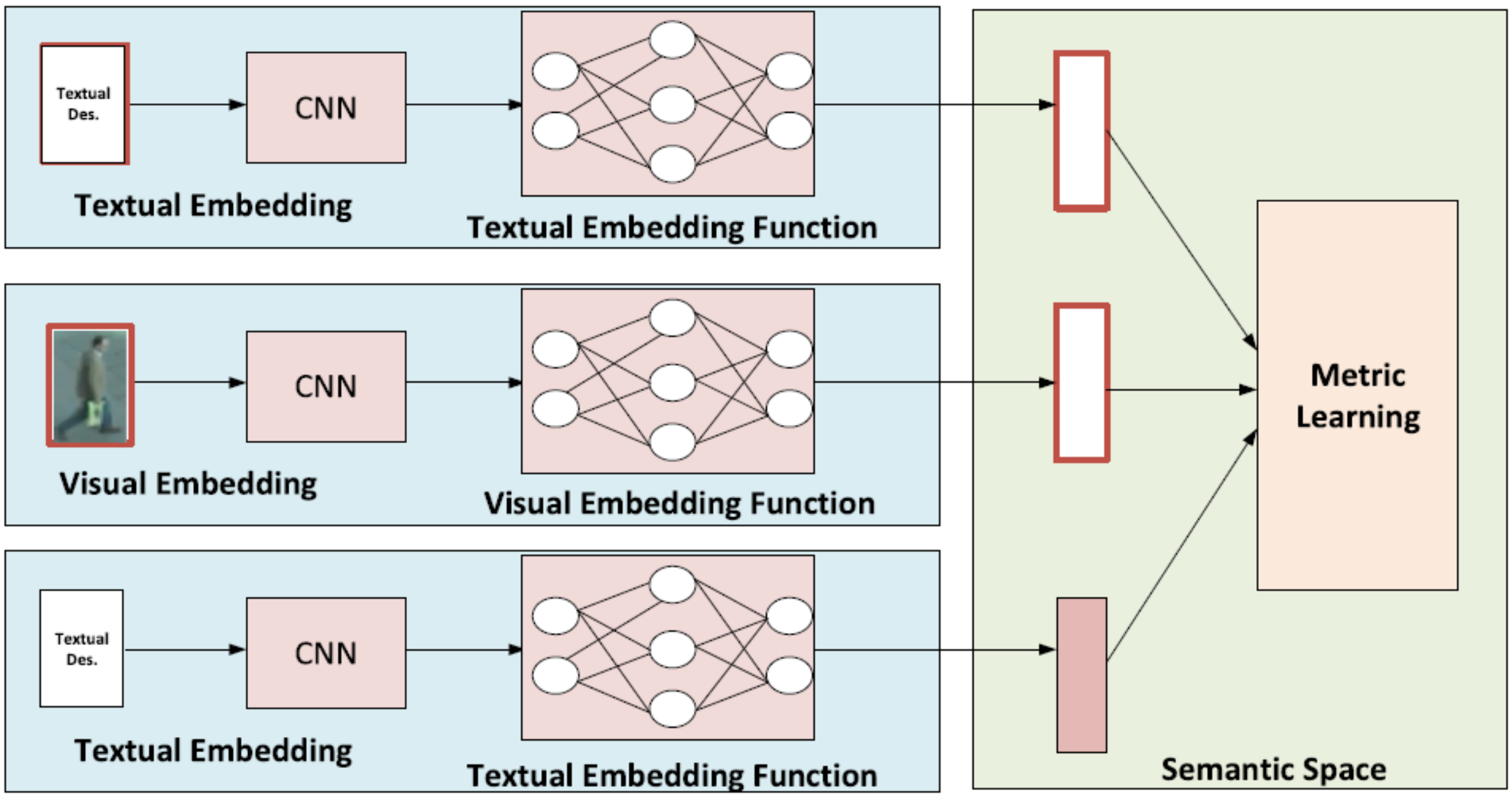

Fig. 13 Triplet Setup to reduce the distance of a positive textual description (top blue box) from an image sample identity (center blue box) and increase the distance from the negative textual description (bottom blue box) [118]

Gait information has proven to be highly discriminative for Re-Id tasks. Gohar et al. [119] formulate a multi-modal Re-Id method performing non-visual gait analysis on information extracted using wearable sensors namely, accelerometer and gyroscope. The proposed method learns discriminative information integrating the temporal aspect of gait data using Recurrent Gated Units (GRU).

The **highlights of modality based deep Re-Id contributions** are:

- Visible RGB images are the most widely used data modality for deep Re-Id methods due to the rich variety of visual information they contain and the growth of several image based Re-Id datasets developed over the years.
- Visible RGB image methods aim to extract discriminative features by using multi-branch networks [110], fusing multiple features together [80], proposing novel loss functions [114] etc.
- Visible RGB image methods face several visual challenges as discussed in Section 4.2 and lose most of their discriminative capabilities in dark/night environments.
- Multi-modal Re-Id methods reduce the dependency on visual information for extracting discriminative features.
- Several multi-modal Re-Id contributions have extracted anthropometric features from depth data [116], modality specific representations from thermal images [117], gait information from accelerometer and gyroscope [119] and combined with visual information to formulate multi-modal representations.

### 4.4 Cross-Domain Re-Id Methods

Based on the kinds of samples present, different re-id datasets hold different generalizations of human appearance. Hence a re-id model trained on one dataset performs poorly on a different dataset. Several works have addressed this issue with domain adaptation techniques attempting to bridge the gap between the learned source domain to the unknown target domain as demonstrated in Table 9. Zhang et al. [58] introduce a novel *Dual Generation Learning* (DGL) method for unsupervised domain adaption such that a re-id model when evaluated on any relevant dataset shows acceptable recognition results. The DGL method generates target style images for samples from source dataset and camera style images for those from target dataset, thereby expanding them to consider varying domain styles. Ganin et al. [121] propose to augment few standard layers and a novel *gradient reversal layer* into deep architectures to learn features that are trained on both labelled source domain as well as unlabelled target domain. Such features cannot discriminate between source and target domain and hence are suitable for domain adaptation. Wang et al. [122] propose a *Deep Multi-Task Transfer Network* (DMTNet) to transfer discriminative features learnt from source domain to target domain by utilizing a cluster estimating algorithm, attribute attention important learning and multi-task learning. Other domain adaptation based re-id works involve refining learned augmented attribute features according to target domain [123], image to image translation using generative adversarial network [124], viewpoint transfer using generative adversarial network [125] etc.

TABLE 9 CROSS-DOMAIN RE-ID CONTRIBUTIONS

| Reference | Year | Cross-Domain Approach |
|---|---|---|
| Xiao et al. [126] | 2016 | Learn generic CNN features embedding from multiple dataset domains and use a novel "Domain Guided Dropout" to mute neurons learning domain specific information thereby improving Re-Id performance |
| Ganin et al. [121] | 2016 | Use few standard layers and a novel "gradient reversal layer" to learn from labelled source domain samples and unlabelled target domain samples |
| Xu et al. [123] | 2019 | Propose a novel "Deep Augmented Attribute Network" (DAAN) to learn augmented feature representations using augmented features and labels from source dataset and refine the learned features to unlabelled target dataset. |
| Zhou et al. [124] | 2019 | Propose a novel "Multi-Camera GAN" (CTGAN) to transfer source dataset samples to multi-camera styles of target dataset |
| Genc et al. [127] | 2019 | Perform domain adaptation by training on different dataset combinations, learning part specific features and learning features form multiple layers and use a CycleGAN to perform camera view adaptations. |
| Wang et al. [122] | 2020 | Propose a novel "Deep Multi-Task Transfer Network" (DMTNet) network for unsupervised cross domain Re-Id including cluster number estimation algorithm, learning of attribute attention importance and transfer of specific multi-task learning across domains |
| Sun et al. [125] | 2020 | Propose a novel "Conditional Transfer Network" (cTransNet) implementing conditional viewpoint transfer using StarGAN and obtain hybrid feature embeddings from original and transformed images to obtain similarity rankings |
| Zhang et al. [58] | 2020 | Propose a novel "Dual Generation Learning" (DGL) method to transfer source dataset images to target style domain and target dataset images to source camera styles to obtain better Re-Id results |

The **highlights of cross-domain Re-Id methods** are:

- Most deep Re-Id methods that are trained/tested on few datasets perform poorly on other datasets.
- Several cross-domain Re-Id approaches have been developed to either learn better generalization across multiple datasets or transfer learned characteristics of source domain to another target domain.
- The cross-domain approaches have achieved sample style transfer from source to target dataset [58], training from multiple dataset combinations [127], transformation of learned features from labelled source to unlabelled target dataset [123], thereby improving the generalizability of deep Re-Id methods.

## 4.5 Metric Learning Methods for Re-Id

Metric learning has proven to be a significant step in computer vision problems such as person re-identification, face recognition etc. Metric learning aims to find a similarity function on extracted features that is used to decrease positive pair distance while increasing negative pair distance [11]. Table 10 highlights the different kinds of metric learning contributions reviewed in this article. Since the underlying data distributions varies with the nature of the computer vision task, the ideal similarity function is mostly task-specific [128]. Traditional metric learning methods focussed on learning linear Mahalanobis based metrics. Such linear metrics utilized the sample distribution centroid and standard deviation to evaluate the similarity with a give sample point. However, such linear metrics often failed to comprehend nonlinear relationships among the samples.

TABLE 10 DEEP METRIC LEARNING METHODS

| Reference | Year | Metric Learning Method | Short Form | Key Idea |
|---|---|---|---|---|
| Hu et al. [128] | 2016 | Deep Transfer Metric Learning | DTML | Cross domain metric learning |
| Lin et al. [129] | 2017 | Generic Similarity Metric | - | Robust to translation and shearing deformation |
| Zhu et al. [130] | 2018 | Deep Hybrid Similarity Learning | DHSL | More discriminative than Euclidean or Cosine distance based similarity |
| Duan et al. [131] | 2018 | Deep Localized Metric Learning | DLML | Metric learning over locally varying data |
| Hu et al. [132] | 2018 | Sharable and Individual Multi-View Deep Metric Learning | MvDML | View invariant metric learning |
| Chen et al. [133] | 2018 | Pose Invariant Deep Metric Learning | PIDML | Pose invariant metric learning |
| Ren et al. [134] | 2019 | Deep Structured Metric Learning | - | Robust person re-identification |
| Ding et al. [135] | 2019 | Robust Discriminative Metric Learning | RDML | Robust to noise |
| Xiong et al. [136] | 2019 | Multiple Deep Metric Learning | - | Instead of feature extraction, utilize a stacked auto-encoder to recognize individuals using multiple similarity probabilities from Softmax regression models |

Recently, deep metric learning techniques have become increasingly popular due to their ability to capture hierarchal nonlinear affiliations. Hu et al. [128] propose a *deep transfer metric learning* (DTML) method that transfers discriminative information from labelled source domain to unlabelled target domain. DTML learns hierarchical nonlinear transformations, maximizing the inter-class disparity, minimizing the intra-class differences and restricting the divergence of source and target domain. Ding et al. [135] increase the generalizability of metric learning by introducing *robust discriminative metric learning* (RDML) which is insensitive to noise unlike most metric learning techniques. A fast low rank model is also used to discover global structure within data and ensure scalability to larger datasets. Lin et al. [129] argue that similarity transformations cannot capture deformations like translation and shearing in cross domain visual matching tasks. The authors propose *a generic similarity metric* that generalizes the similarity transformation to affine transformation capable of capturing complex deformations. While most metric learning techniques focus on the overall similarity learning of samples, Duan et al. [131] propose the *deep localised metric learning* (DLML) to learn multiple metrics fine-grained over numerous local subspaces. DLML specializes in handling data varying locally. Hu et al. [132] introduce the *sharable and individual multi-view deep metric learning* (MvDML) to utilize multi-view data in best possible manner. MvDML learns the best combination of distance metrics by focussing on both individual view specifics metric as well as a combined multi-view representation. Chen et al. [133] propose a novel *pose invariant deep metric learning* (PIDML) method that utilizes pose invariant embedding [93] and an improved triplet loss to achieve pose invariance for metric learning. Ren et al. [134] propose a *deep structured metric learning method* that utilizes a novel structured loss function to achieve robust person re-identification. The proposed loss skips positive sample pairs of small distance and negative sample pairs having large distance.

The **highlights of deep metric learning based Re-Id methods** are:

- Metric learning means to learn a similarity function that pulls features from same identity samples closer and pushes different identity features away.
- Deep learning based metric learning have gained popularity in recent years due to their ability to learn nonlinear sample associations unlike the traditional Mahalanobis metric learning only linear relationships.
- Several novel deep metric learning approaches have been proposed that give better discrimination capability than the usual Euclidean or Cosine measure [130], can perform cross-domain learning [128], and are robust to noise [135], deformation [129] etc.

## 5. Video Based Deep Re-Id Contributions

This section explores the video based deep person re-identification methods.

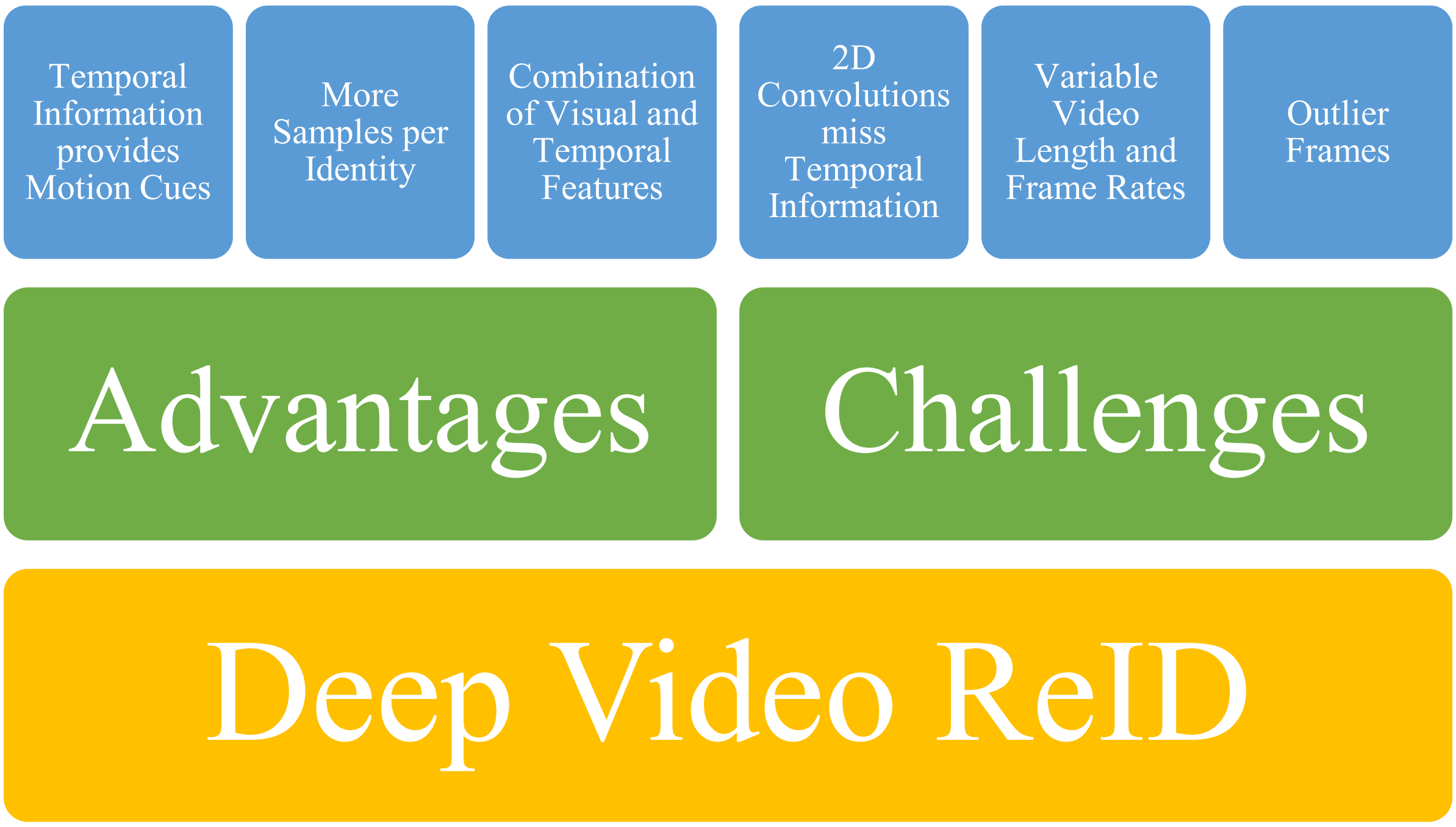

Fig. 14 Advantages and Challenges of Video based Person Re-Identification methods

Video Re-Id methods received less attention compared to image based techniques in the initial years of Re-Id research [67]. A major reason was the lack of large-scale video datasets. However, recent years have witnessed the growth of several video based Re-Id datasets such as [32], [36] and [38] encouraging the development of video Re-Id methods. Compared to images, video sequences have advantageous characteristics that can be exploited for Re-Id tasks. *Firstly*, a video sequence comprising of multiple frames holds essential temporal information across frames giving crucial motion cues which are absent from image samples. *Secondly*, multiple frames from each video provide numerous visual examples per identity adding diversity in samples. *Thirdly*, while image based methods mostly focus on exploiting visual features, video Re-Id methods target a combination of visual and temporal features to extract discriminative features for Re-Id tasks which prove to be more robust to Re-Id challenges. While video based methods give rise to more discriminative feature embeddings, they also add to the existing Re-Id challenges as demonstrated in Fig. 14. *Firstly*, since the simplest way of generating video features is by fusing frame level features together and frame level features are based on 2D convolution operations that totally neglect the temporal aspect of frame sequences, the fusion operation misses all of the temporal cues crucial for video Re-Id. *Secondly*, videos have different frame rates and different time series that makes it hard to make comparisons among samples. *Thirdly*, not all frames provide discriminative information and some outlier frames prove misleading in learning robust video representations.

Recent developments in Recurrent Neural Networks (RNN) and Long Short-Term Memory (LSTM) models [65] have boosted video Re-Id by providing the capability of extracting motion cues from temporal information for robust video representations. McLaughlin et al. [137] propose a video re-identification system based on recurrent neural network for wide area tracking. The proposed network uses a recurrent layer to combine frame-level details from all frames into a single combined appearance feature representing the entire video showing competent recognition results. Attention modules have played key role in video Re-Id to identify and isolate outlier frames. They also help in extracting discriminative regions within video frames. Wu et al. [67] propose a Siamese attention network that learns to realize which regions (where) from which frames (when) are relevant for comparing identities. The attention mechanism learns the most relevant features by focussing on distinct regions helping to identify given identity. Zhang et al. [37] propose a *self and collaborative attention network* (SCAN) for video re-id. The proposed model takes a pair of videos as input, aligns and compares their discriminative frames using a generalized similarity measurement module and refines intra-sequence and inter-sequence features from videos using a non-parametric attention module. Wu et al. [138] argue that merely combining frame-wise features to obtain overall video features is ineffective as the temporal cues get lost in 2D convolutions. Authors propose a novel 3D Person VLAD Aggregation layer that helps to extract appearance and motion characteristics of entire video. The model also handles occlusions and misalignments through soft attention modules. Zhang et al. [74] propose a *multi-scale spatial-temporal attention network* (MSTA) that focusses attention to regions within video frames at different scales. It contains a ResNet50 based encoder responsible for extracting frame-level features from discriminative regions and an aggregator to fuse features from different scales Fig. 15.

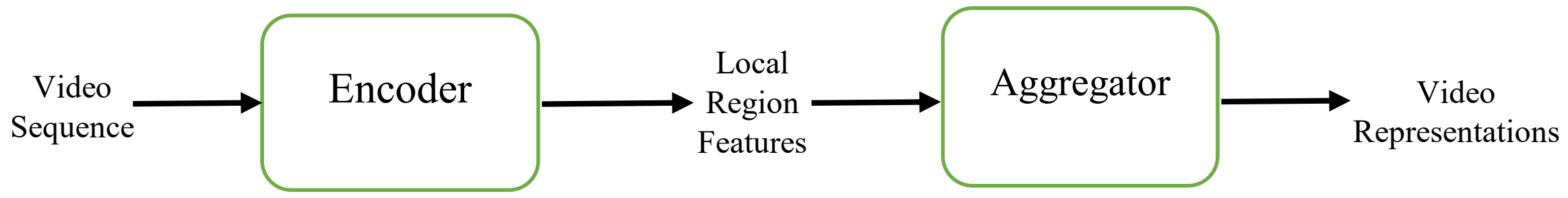

Fig. 15 Architecture of proposed MSTA model. The encoder extracts framewise features and the aggregator fuses the frame-level features to obtain video representations *[74]*

Table 11 dives into several contributions highlighting the diverse approaches towards video person re-identification using deep learning.

TABLE 11 VIDEO BASED DEEP RE-ID CONTRIBUTIONS

| Reference | Year | Key Idea | Dataset | Rank 1 Result (%) |
|---|---|---|---|---|
| Sun et al. [139] | 2018 | Extract visual features using a "Two-Branch Appearance Feature sub-structure" (TAF) and temporal features using a "Optical Flow Temporal Feature sub-structure" (OTF) and use a pair of Siamese networks to learn similarity measure among pairwise visual and temporal features. A saliency learning fusion layer learns fusion of global and local appearances. | MARS | 73.00 |
| | | | iLID-VIDS | 59.00 |
| | | | PRID2011 | 79.00 |
| Wang et al. [140] | 2018 | Perform image-to-video Re-Id using a novel end-to-end "Point-to-Set Network" (P2SNet) that takes both image and video as input and jointly learns their features using point-to-set distance metric. A *k*NN-triplet module helps to focus only on relevant video frames | MARS | 55.25 |
| | | | iLID-VIDS | 73.31 |
| | | | PRID2011 | 40.00 |
| Meng et al. [141] | 2019 | Learn view-specific and feature-specific transformations using a novel "Deep Asymmetric Metric Learning" from a two stream neural net to counter view-specific bias and feature-specific bias in appearance and motion features due to variations in view, lightning, background clutter etc | MARS | 74.65 |
| | | | iLID-VIDS | 77.33 |
| | | | PRID2011 | 87.00 |
| Wu et al. [67] | 2019 | Propose a novel "Siamese attention architecture" to jointly learn feature representation and similarity measure for video Re-Id using an attention mechanism focussing on relevant frames (when) and regions of interest within frames (where) | MARS | 86.60 |
| | | | iLID-VIDS | 86.50 |
| | | | PRID2011 | 98.80 |
| Ksibi et al. [33] | 2019 | Propose a "Deep Spatio-Temporal Appearance" (DSTA) descriptor that uses a "Deep Salient-Gaussian Wighted Fisher Vector" (SGFV) to exploit trajectory information to handle misalignment of person tracklets and eliminate background noise using Gaussian and Saliency maps | MARS | 76.70 |
| | | | iLID-VIDS | 80.00 |
| | | | PRID2011 | 92.70 |
| Zhang et al. [37] | 2019 | Propose a novel "Self-and-Collaborative Attention Network" (SCAN) that utilizes two attention subnetworks (self-attention subnetwork and collaborative attention subnetwork) to select features from informative video frames and align discriminative frames from probe and gallery frames respectively and finally use a generalized similarity measure to compare video pair representations | MARS | 87.20 |
| | | | iLID-VIDS | 88.00 |
| | | | PRID2011 | 95.30 |
| Wu et al. [138] | 2019 | Utilize novel "3D Person VLAD Aggregation layer" based on the vector of locally aggregated descriptors to capture motion based features along with appearances which is usually missed by 2D ConvNets and learn "global representations" for a full length video robust to occlusion and misalignment using soft attention module to learn 3D part alignment | MARS | 80.80 |
| | | | iLID-VIDS | 69.40 |
| | | | PRID2011 | 87.60 |
| Liu et al. [142] | 2019 | Propose a novel "Dense 3D Convolutional Network" (D3DNet) that uses numerous three dimensional dense blocks and transition layers to extract discriminative features from appearance and temporal domains capturing visual and motion cues (both short term and long term) from videos without additional modules. Implement a combination of identification and center loss to reduce intra-class disparity and increase inter-class disparity | MARS | 76.00 |
| | | | iLID-VIDS | 65.40 |
| McLaughlin et al. [137] | 2019 | Extract frame-wise CNN features and use a temporal pooling recurrent layer to combine all time step features and obtain feature representation for entire video sequence | iLID-VIDS | 58.00 |
| | | | PRID2011 | 70.00 |
| Zhang et al. [74] | 2020 | Propose a novel "Multi-Scale Spatial-Temporal Attention" (MSTA) model that pays attention to different regions within each video frame at different scales to incorporate essential regions into whole video spatio-temporal representations. MSTA contains an encoder to extract frame wise features and an aggregator to fuse features | MARS | 82.28 |
| | | | iLID-VIDS | 70.10 |
| | | | PRID2011 | 91.20 |
| Wu et al. [143] | 2020 | Use variational recurrent neural networks (VRNNs) to conduct a deep few-shot adversarial learning to extract discriminative features that are view invariant | MARS | 54.60 |
| | | | iLID-VIDS | 60.10 |
| | | | PRID2011 | 79.20 |
| Avola et al. [144] | 2020 | Propose a novel "LSTM based Re-Id Hashing model" that exploits bone proportion, gait and movement features of 2D skeletons extracted from RGB video frames. LSTM is used to learn temporal correlation between different frames while two dense layers are responsible for implementing bodyprint hashing via binary coding capable of unbounded labelling of all individuals possible | MARS | 86.50 |
| | | | iLID-VIDS | 73.40 |
| | | | PRID2011 | 82.70 |
| Wu et al. [145] | 2020 | Propose a novel "Adaptive Graph Representation Learning" method that uses adaptive structure-aware adjacency graphs via graph neural networks (GNN) highlighting two kinds of relations: pose alignment connection to capture human part relations and feature affinity connection to model semantic relationship among | MARS | 89.80 |
| | | | iLID-VIDS | 83.70 |
| | | | PRID2011 | 93.10 |

| Reference | Year | Key Idea | Dataset | Rank 1 Result (%) |
|---|---|---|---|---|
| | | features from various regions across frames. A novel regularization is used to capture temporal resolution invariant features for the entire video sequence | DukeMTMC VideoRe-Id | 96.70 |
| Jiang et al. [59] | 2020 | Propose a novel framework "Spatial Transformed Partial Network" (STPN) that aligns frames to extract robust regional features. A novel "Weighted Triple-Sequence Loss" (WTSL) is used to exclude outlier frames in video-level features. Model is jointly optimized for frame-level and video level features. | MARS | 85.90 |
| | | | iLID-VIDS | 82.20 |
| | | | PRID2011 | 95.20 |

## 6. Conclusion and Future Directions

This review provides a comprehensive and exhaustive analysis of deep learning based person re-identification methods. The objective of this work is to give the readers a thorough understanding of different approaches towards deep Re-Id. This reviewed literature has been divided into several logical categories as demonstrated in the taxonomy diagram Fig. 2. These approaches have been classified on the basis of adopted architecture types (classification, verification, triplet based, part based and attention models) integrating different kinds of losses (softmax, triplet), the common Re-Id challenges faced (variations in pose, lightning, view, scale, partial or complete occlusion, background clutter), image based methods and multi-modal Re-Id methods reducing the dependency on visible RGB approaches, cross-domain methods to improve generalizability of approaches across different datasets, metric learning approaches to learn ideal similarity functions and deep video Re-Id methods exploiting both spatial and temporal cues from multiple frames of a video sequence. Each category presented as a separate section provides an extensive look into these contribution types and the highlights at the end of each section gives a quick overview of the reviewed methods. Tables 3-11 provide the *key ideas* behind numerous deep Re-Id methods across various categories.

Part-Based and Attention architectures are the more popular architecture types in recent times owning to their ability to find regions of rich information and extraction of finer visual cues. Re-Id challenges have been conquered using different techniques like using pose estimation to achieve pose invariance [93], obtain scale invariance through multi-scale input convolutions [61] etc. Some multi-modal approaches have helped to reduce the over dependence on visible RGB based methods making them more viable for darker environments with fewer visual cues [117]. Deep video Re-Id methods provide more discriminative information when compared to image based approaches due the motion cues present in temporal information across frames. However, video Re-Id presents its own challenges of finding discriminative regions both spatially (within frames) and temporally (across frames) as demonstrated in Fig. 14.

Based on this review of deep Re-Id methods, the following research gaps can guide the future research motivations:

- Several Re-Id datasets have been collected under controlled environment like research labs, college campuses etc. While numerous Re-Id approaches have attained high performance on these datasets, their results suffer tremendously in realistic scenes. Datasets containing more real-like scenarios like [29] are needed to enhance the potential of proposed Re-Id methods.

- Very few multi-modal datasets exist which seriously limit the growth of multi-modal Re-Id approaches. Preparation of large scale multi-modal datasets can greatly contribute to deep Re-Id research.
- Gohar et al. [119] use gait-data collected from wearable sensors fixed on the chest of test subjects. Multi-modal methods can be developed that can process data from sensors without considering the positioning or orientation of data recording devices.
- There are few end-to-end Re-Id research contributions that involve both person detection and re-identification together in a single framework. Since most datasets are collected under controlled environment, the person detection is usually performed automatically. End-to-end Re-Id is a promising research direction.
- GAN based Re-Id works have greatly supported the style transfer requirements of cross-domain Re-Id approaches. However, the low to medium quality of generated samples has limited the performance of these approaches. An improvement in sample generation quality can significantly boost cross-domain Re-Id approaches.
- Most part-based Re-Id contributions have focussed on systematic comparison between corresponding part regions of input pairs. But the contextual relationship among different regions is mostly ignored. Preserving the semantic relationship among different parts like [89] is a potential way of improving part-based methods further.
- Attribute-based methods have become increasingly popular in finding finer visual cues. These attribute-based methods can be extended further for applications like part localization like in [88] to achieve better Re-Id performance.